\pdfoutput=1

\documentclass[11pt]{article}
\usepackage{float}

\usepackage[]{acl2023}

\usepackage[utf8]{inputenc}
\usepackage{pgfplots}
\usepackage{dsfont}
\DeclareUnicodeCharacter{2212}{−}
\usepgfplotslibrary{groupplots,dateplot}
\usetikzlibrary{patterns,shapes.arrows}
\pgfplotsset{compat=newest}

\usepackage{tikzscale}
\usepackage{amsmath}
\newcommand{\probP}{\text{I\kern-0.15em P}}
\newcommand{\specialcell}[2][c]{%
  \begin{tabular}[#1]{@{}c@{}}#2\end{tabular}}

\newcommand{\specialcellleft}[2][l]{%
\begin{tabular}[#1]{@{}l@{}}#2\end{tabular}}

\usepackage{multirow, colortbl}

\usepackage{tabularx,booktabs}
\usepackage{makecell}

\usepackage[normalem]{ulem}
\useunder{\uline}{\ul}{}

\usepackage{graphicx}

\definecolor{ablation6}{HTML}{fcefed}
\definecolor{ablation_tie}{HTML}{fce3e1}

\definecolor{ablation5}{HTML}{fcd8d4}
\definecolor{ablation4}{HTML}{FBC3BC}
\definecolor{ablation3}{HTML}{F7A399}
\definecolor{ablation2}{HTML}{F38375}
\definecolor{ablation1}{HTML}{EF6351}
\definecolor{UMDred}{HTML}{ed1c24}
\definecolor{UMDyellow}{HTML}{ffc20e}
\definecolor{CustomGreen}{HTML}{1FC801}

\usepackage[]{algpseudocode}
\usepackage[]{algorithm}
\usepackage{float}
\algtext*{EndFor}%
\algtext*{EndProcedure}%

\usepackage{booktabs}
\usepackage[normalem]{ulem}
\useunder{\uline}{\ul}{}

\usepackage{times}
\usepackage{latexsym}
\usepackage{adjustbox}

\usepackage[T1]{fontenc}
\usepackage{amsmath}
\usepackage{amssymb}
\usepackage{booktabs}
\usepackage{multirow}
\usepackage{scalerel,xparse}
\usepackage[utf8]{inputenc}

\usepackage{cleveref}
\usepackage{microtype}
\usepackage{enumitem}
\crefformat{section}{\S#2#1#3}
\crefformat{subsection}{\S#2#1#3}
\crefformat{subsubsection}{\S#2#1#3}

\title{It's Not Easy Being Wrong:\\Large Language Models Struggle with Process of Elimination Reasoning}

 \author{Nishant Balepur $\quad$ Shramay Palta  $\quad$ Rachel Rudinger \\
 University of Maryland, College Park, USA \\
 \texttt{\{nbalepur, spalta, rudinger\}@umd.edu}}

\begin{document}
\maketitle

\begin{abstract} {
Chain-of-thought (COT) prompting can help large language models (LLMs) reason toward \emph{correct} answers, but its efficacy in reasoning toward \emph{incorrect} answers is unexplored. This process of elimination (PoE), when used with COT, can enhance self-consistency, interpretability, and tasks such as medical diagnoses of exclusion. Thus, we propose PoE with COT, where LLMs must reason toward incorrect options on multiple-choice questions. We evaluate the ability of GPT-3.5, LLaMA-2, and Falcon to perform PoE with COT on a total of four commonsense and scientific reasoning datasets. We find that the strategy of PoE always underperforms the strategy of choosing the correct answer. The agreement of these strategies is also lower than the self-consistency of each strategy. To study these issues further, we conduct error analyses and give suggestions for future work.\footnote{Our code is available at: \url{https://github.com/nbalepur/PoE}}
}
\end{abstract}


\section{Introduction}

Recent research has aimed to unlock the reasoning capabilities of large language models (LLMs) \cite{nye2022show}. As part of this effort, researchers have proposed techniques such as chain-of-thought (COT) prompting to help LLMs verbally reason toward correct answers \cite{wei2022chain, Kojima2022LargeLM}. Such reasoning can improve the accuracy and interpretability of LLM decision-making \cite{creswell2023selectioninference, huang-chang-2023-towards}. 

\begin{figure}
    \centering
    \includegraphics[width=\linewidth]{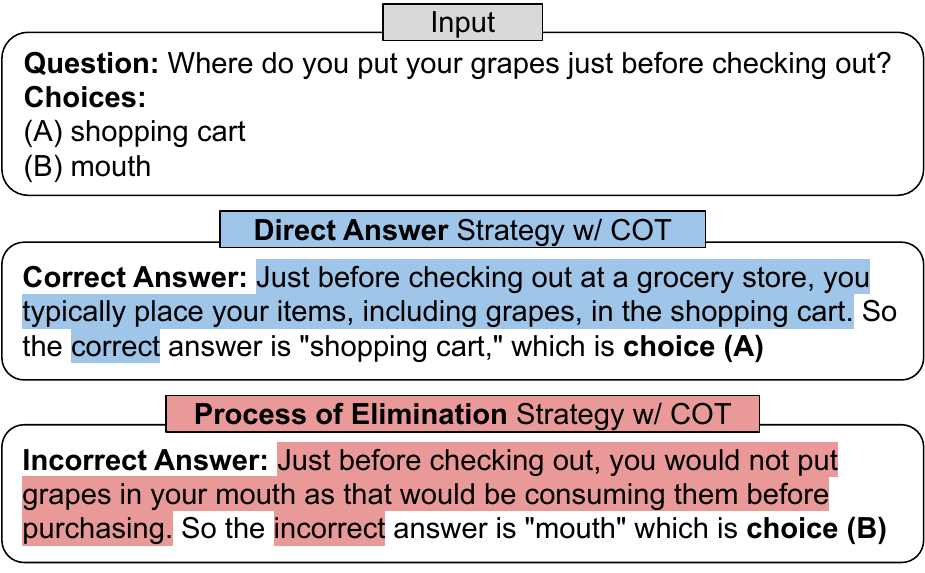}
    \caption{ChatGPT using direct answer and process of elimination strategies via chain-of-thought prompting.}
    \label{fig:intro}
\end{figure}

While several works use COT to select \emph{correct} answers (Figure~\ref{fig:intro} \textbf{\textcolor{blue}{blue}}), they do not study if COT can identify \emph{incorrect} answers (Figure~\ref{fig:intro} \textbf{\textcolor{red}{red}}). Thus, drawing from test-taking strategies for multiple-choice (MC) exams \cite{tversky1972choice}, we propose a new task to probe LLM reasoning: \textbf{process of elimination (PoE) with COT}. Previous works have performed PoE by discarding low-confidence options when selecting correct answers \cite{ma2023poe}, but whether generative LLMs can directly reason toward incorrect options with COT is unexplored.

Analyzing PoE with COT poses several benefits. \textbf{First}, it can give insights into the consistency of LLM reasoning \cite{wang2023selfconsistency}. For 2-choice MC questions, we can study if the decisions from PoE with COT and directly answering the question with COT agree, as both should arrive at the same answer (\cref{subsection:consistent}). \textbf{Second}, PoE with COT can enhance LLM interpretability. COT reasoning is often seen as a rationale as to why the model thinks an option is correct, but users may also want to know why the model thinks alternatives are incorrect. \textbf{Third}, many applications benefit from PoE, such as reaching medical diagnoses via exclusion \cite{fred2013diagnosis, kline2018diagnosis}, troubleshooting by eliminating root causes \cite{gutoff2006coating, gugerty2007cognitive}, and ruling out scientific hypotheses using new evidence \cite{norton1995eliminative, forber2011reconceiving}. Thus, to enable interpretable LLMs in domains like medicine, customer service, and research, we must know: \emph{Do LLMs have the ability to perform PoE with COT?}

To study this question, we independently prompt the GPT-3.5 \cite{10.5555/3495724.3495883}, LLaMA-2 \cite{touvron2023llama}, and Falcon \cite{penedo2023refinedweb} LLMs to directly select the correct option and eliminate the incorrect one, with and without COT, on 2-choice commonsense \cite{sap-etal-2019-social, talmor-etal-2019-commonsenseqa} and scientific \cite{clark2018think, mihaylov-etal-2018-suit} reasoning datasets. In this 2-choice setup, we can juxtapose LLM reasoning abilities in direct answer selection against PoE, as ideal LLMs would always have both strategies agree.

We find that PoE \textbf{always} underperforms direct answer selection for our LLMs, but this gap narrows as model size scales, implying that PoE with COT may only be attainable for larger LLMs (\cref{subsection:accuracy}). Further, these two strategies do not reliably arrive at the same answer, revealing an inherent inconsistency in LLM reasoning (\cref{subsection:consistent}). For more insights, we analyze the errors in the rationales of PoE with COT, finding that most stem from reasoning errors and task misalignment (\cref{subsection:error_analysis}), and show how these 2-choice errors propagate when using PoE on full MC questions (\cref{subsection:iterative_poe}). Based on our results, we give suggestions for future work. Our contributions are:

\noindent \textbf{1)} We introduce PoE with COT, a new reasoning task for LLMs that can benefit interpretability, self-consistency, and downstream applications. \\
\noindent \textbf{2)} We benchmark the abilities of GPT, LLaMA-2, and Falcon to perform PoE with and without COT on scientific and commonsense reasoning datasets. \\
\noindent \textbf{3)} We evaluate the consistency and errors of PoE with COT to suggest directions for future research.


\section{Problem Definition}

We study selecting correct and incorrect options in MC settings, where we are given a question $q$ and $n$ choices $\mathcal{C}$, exactly one of which is correct. To simplify our discussion, we define two strategies: \\
\noindent \textbf{1) Direct Answer (DA) Strategy:} The LLM aims to select the correct answer choice $c_t \in \mathcal{C}$.\\
\noindent \textbf{2) Process of Elimination (PoE) Strategy:} The LLM aims to select an incorrect choice $c_f \in \mathcal{C}$.

We ideally want to perform PoE iteratively, generalizing the strategy for any number of choices.
But as a prerequisite, we must know: Are LLMs inherently capable of reasoning toward incorrect choices?
To answer this question, we analyze the logical consistency of LLMs via 2-choice questions ($\mathcal{C} = \{c_t, c_f\}$), as a robust LLM would always have DA and PoE agree: predicting that one choice is correct entails that the other is incorrect. 
If PoE fails with two choices, it reveals that LLMs are inherently weaker at picking incorrect options and these errors will propagate when using PoE iteratively.
We first show that PoE is unreliable in 2-choice settings (\cref{subsection:accuracy}), and later show this leads to error propagation in iterative settings (\cref{subsection:iterative_poe}).





\section{Experimental Setup} \label{section:experimental_setup}

\subsection{Strategy Implementation}

To provide sufficient context for the LLMs, we use few-shot (10) prompts to implement the strategies, leaving the analysis of 0-shot PoE with COT for future work. We construct the following prompts: \\
\noindent \textbf{1) DA Base} and \textbf{PoE Base:} These prompts task the LLM with performing the DA and PoE strategies without reasoning. The in-context examples in these prompts follow a similar format to Figure~\ref{fig:intro}, but only the correct/incorrect answer choice ((A) or (B)) follow the ``Correct/Incorrect Answer:'' labels. \\
\noindent \textbf{2) DA COT} and \textbf{PoE COT:} These prompts add COT to the DA and PoE Base prompts. The few-shot examples follow the same format as Figure~\ref{fig:intro}, requiring the LLM to give a step-by-step rationale before answering. Notably, the DA and PoE COT examples are created to be \textbf{distinct} such that DA/PoE reason toward the correct/incorrect answer without discussing the other choice's validity. This is meant to prevent unhelpful PoE rationales like ``The answer is (A). So the incorrect answer is (B).''

To study if COT reasoning improves the accuracy of each strategy (\cref{subsection:accuracy}), we can compare the accuracy of the Base prompts with their COT counterparts (e.g. PoE Base vs PoE COT). Further, comparing the outcomes of the DA and PoE prompts (e.g. DA COT vs PoE COT) allows us to measure the LLMs' logical consistency in choosing correct answers and eliminating incorrect options (\cref{subsection:consistent}). 

The COT rationales are written by one Ph.D. student and verified by a second (the authors). We also prepend a natural language instruction to the prompt explaining the strategy. More prompting details and examples can be found in Appendix~\ref{appendidx:prompts}.


\begin{figure*}
    \centering
    \includegraphics[width=\linewidth]{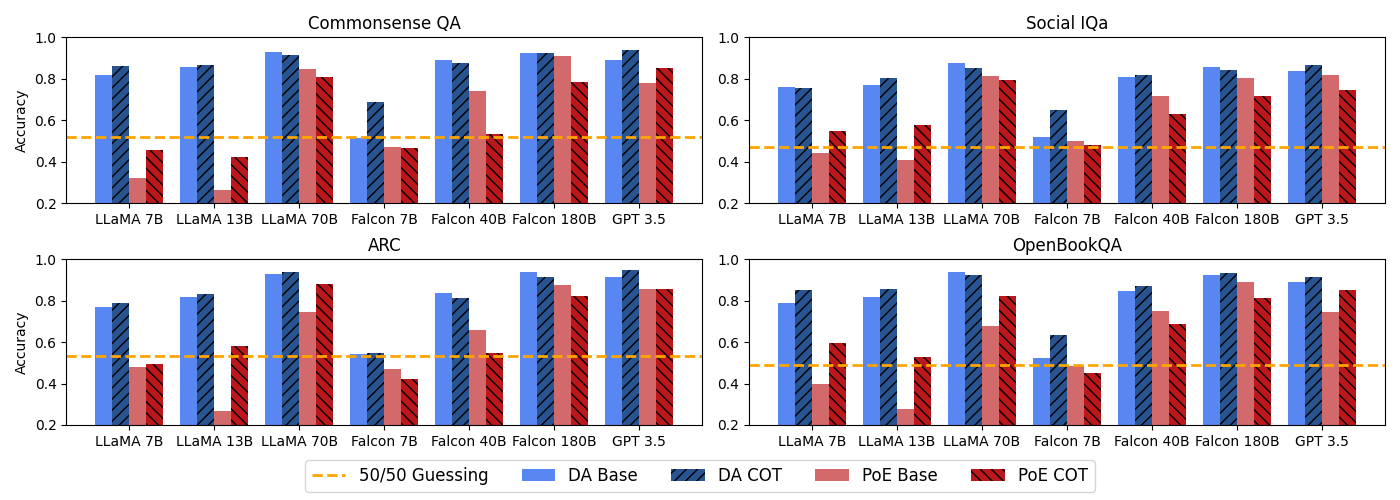}
        \caption{Accuracy of Direct Answer and Process of Elimination, with and without chain-of-thought, on commonsense (CQA, SIQA) and scientific (ARC, OpenBookQA) reasoning datasets. Numerical results are in Appendix~\ref{appendix:numerical_eval}.}
    \label{fig:benchmark}
\end{figure*}

\subsection{Datasets}

We examine multiple-choice commonsense and scientific reasoning questions, given that the DA strategy has shown to perform better on these questions using COT reasoning \cite{wei2022chain, NEURIPS2022_11332b6b, Zhang2023MultimodalCR, Lanham2023MeasuringFI}.\footnote{We do not use binary Yes/No datasets, like BoolQA \cite{clark2019boolq}, as the reason for why a correct answer is correct and an incorrect answer is incorrect are the exact same. Thus, they do not align with our requirement of distinct rationales.}

\textbf{Commonsense reasoning} questions have models reason about knowledge understood by most humans. We use Social IQa \cite[SIQA]{sap-etal-2019-social}, containing questions about social interactions, and CommonsenseQA \cite[CQA]{talmor-etal-2019-commonsenseqa}, concerning knowledge encoded in ConceptNet. 

\textbf{Scientific reasoning} datasets have models make logical inferences based on scientific facts. We use ARC \cite{clark2018think} and OpenBookQA \cite[OBQA]{mihaylov-etal-2018-suit}, testing factual recall, multi-step reasoning, and some commonsense. 

We sample 500 random questions from the test set of each dataset for evaluation, and sample 10 random training examples for the 10-shot prompts. To leave each question with only two choices, we randomly eliminate all but one incorrect choice. Dataset statistics can be found in Appendix~\ref{appendidx:dataset}.


\subsection{Models}

We study three families of LLMs. First, we use GPT-3.5 (\texttt{gpt-3.5-turbo-0613}) \cite{NEURIPS2022_b1efde53} with the OpenAI API. We also use two open-source base LLMs: LLaMA-2 (7B, 13B, 70B) \cite{touvron2023llama} and Falcon (7B, 40B, 4-bit 180B) \cite{penedo2023refinedweb}, loaded via huggingface. Each LLM decodes with 0.3 temperature.


\section{Results}

\subsection{Can LLMs Perform PoE?} \label{subsection:accuracy}

In Figure~\ref{fig:benchmark}, DA Base and DA COT surpass their PoE counterparts for \textbf{every} model and dataset. Notably, smaller LLMs (7B, 13B) using PoE often underperform a 50/50 guessing model, suggesting that the smaller models are fully misunderstanding the PoE task, despite being given 10 examples. In contrast, the accuracy discrepancy between the PoE and DA strategies is less pronounced for larger LLMs, implying that PoE may only be an attainable ability for larger LLMs \cite{wei2022emergent}. Our results show that LLMs, especially smaller models, may have a bias toward choosing correct answers, and thus struggle to perform PoE accurately.
We speculate this occurs because LLM pre-training data likely contains many MCQA questions and explanations justifying the correct choice, but far fewer explanations justifying incorrect choices, leading to a bias towards choosing correct answers.

Further, while COT tends to slightly improve the accuracy of the DA strategy across LLMs, this is not true for PoE. Specifically, DA COT has equal or better accuracy than DA Base in 19/24 cases (8/12 for LLaMA, 7/12 for Falcon, 4/4 for GPT). However, while PoE COT surpasses PoE Base in 10/12 cases for LLaMA and 2/4 cases for GPT, this improvement is never seen for Falcon, indicating a weakness in reasoning. This discrepancy across models further motivates PoE as a promising task. While COT tends to improve or maintain accuracy when picking correct answers, its accuracy wavers by LLM when eliminating choices. Thus, we believe that the accuracy of PoE with COT could be used to evaluate the reasoning abilities of LLMs.

\subsection{Are PoE and DA Consistent?} \label{subsection:consistent}

In Table~\ref{table:consistency}, we study the logical consistency of DA and PoE, i.e., if the strategies converge to the same answer. For \textbf{every} LLM and dataset, we find that DA more often agrees with itself upon a repeated inference (i.e. self-consistency) than agrees with its PoE counterpart. This suggests that LLMs have an innate logical inconsistency when asked to execute these two strategies, which cannot be ascribed to sampling variation from our temperature selection. 

We suggest three future directions based on our results. \textbf{First}, the agreement of DA and PoE can assess logical robustness. Future LLMs can aim to achieve DA/PoE consistency closer to DA self-consistency. \textbf{Second}, measuring when DA and PoE agree could be used for LLM confidence calibration \cite{cheng2023prompting}, as an LLM that arrives at the same solution with diverse strategies may be more confident. \textbf{Lastly}, as LLMs have shown to improve rationales by combining multiple reasoning chains \cite{Yoran2023AnsweringQB}, future works could similarly try to synthesize DA and PoE reasoning chains. 

\begin{table}[t]
\setlength{\tabcolsep}{3.5pt}
\centering
\small
\begin{tabular}{@{}ccccccccc@{}}
 & \multicolumn{4}{c}{\emph{Falcon 180B}} & \multicolumn{4}{c}{\emph{GPT-3.5}} \\ \midrule
\multicolumn{1}{c|}{\multirow{2}{*}{Dataset}} & \multicolumn{2}{c|}{\textbf{DA Base}} & \multicolumn{2}{c|}{\textbf{DA COT}}  & \multicolumn{2}{c|}{\textbf{DA Base}} & \multicolumn{2}{c}{\textbf{DA COT}}  \\
\multicolumn{1}{c|}{}  & PoE & \multicolumn{1}{c|}{Self}  & PoE & \multicolumn{1}{c|}{Self}  & PoE & \multicolumn{1}{c|}{Self}  & PoE & Self  \\ \midrule
\multicolumn{1}{c|}{CQA} & 91.4 & \multicolumn{1}{c|}{\underline{97.6}} & 78.2 & \multicolumn{1}{c|}{\underline{92.9}} & 81.3 & \multicolumn{1}{c|}{\underline{97.3}} & 86.0 & \underline{97.6} \\ 
\multicolumn{1}{c|}{SIQA} & 86.6 & \multicolumn{1}{c|}{\underline{95.0}} & 69.4 & \multicolumn{1}{c|}{\underline{91.4}} & 80.7 & \multicolumn{1}{c|}{\underline{97.0}} & 76.2 & \underline{95.2} \\ \midrule
\multicolumn{1}{c|}{ARC} & 89.3 & \multicolumn{1}{c|}{\underline{98.5}} & 79.1 & \multicolumn{1}{c|}{\underline{90.8}} & 86.6 & \multicolumn{1}{c|}{\underline{97.5}} & 85.5 & \underline{97.0} \\ 
\multicolumn{1}{c|}{OBQA} & 89.7 & \multicolumn{1}{c|}{\underline{96.2}} & 77.2 & \multicolumn{1}{c|}{\underline{95.1}} & 72.3 & \multicolumn{1}{c|}{\underline{98.2}} & 81.4 & \underline{96.8} \\ \bottomrule
\end{tabular}
\caption{\label{table:consistency}Agreement of DA strategies (Base/COT)~with \textbf{PoE} counterparts vs. \textbf{Self}-consistency. More consistent methods are \underline{underlined}. Full results in Appendix~\ref{appendidx:consistency}.}
\end{table}

\begin{figure}[t]
    \centering
    \includegraphics[width=\linewidth]{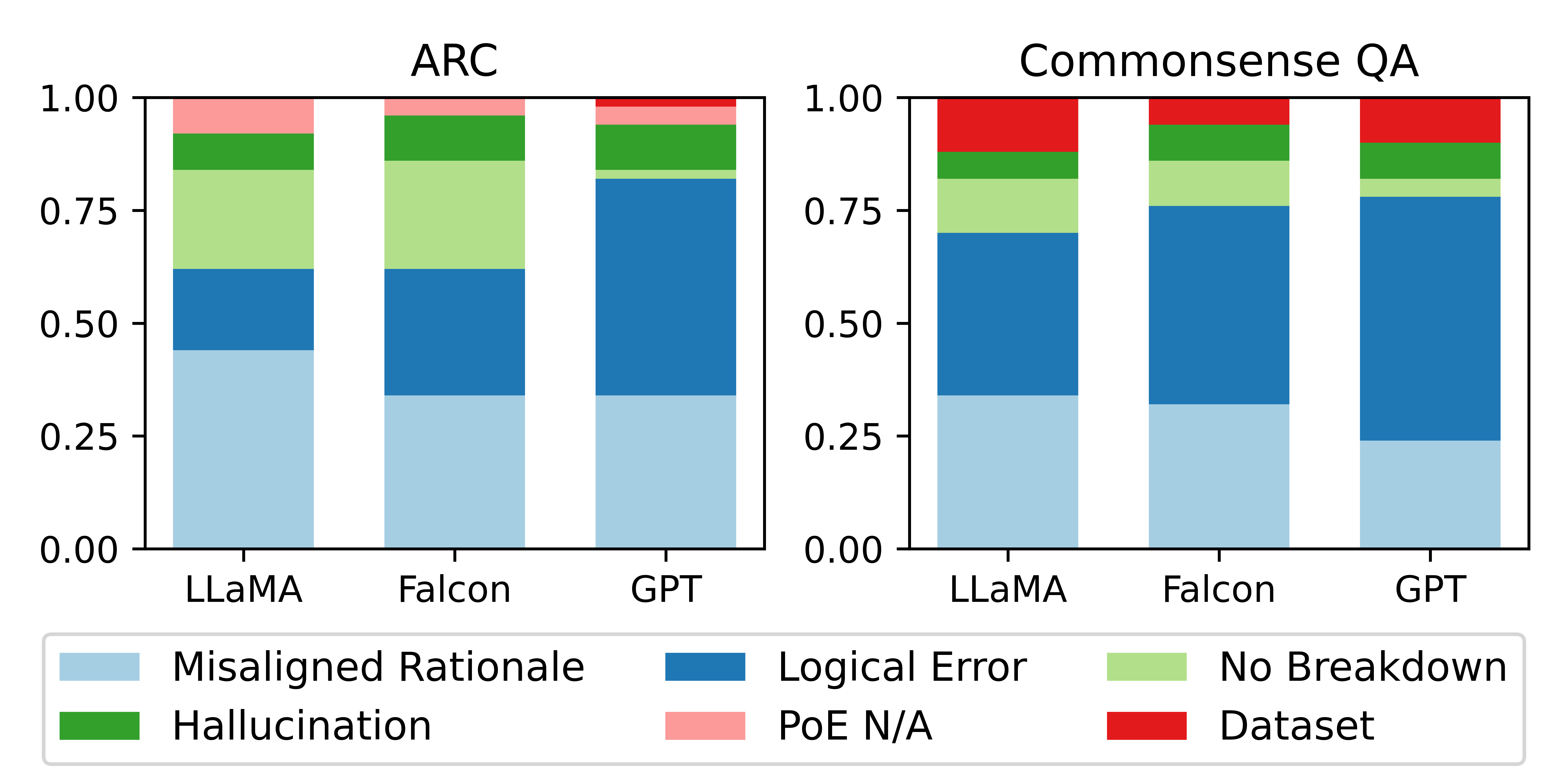}
    \caption{Error distribution of PoE COT on ARC/CQA.}
    \label{fig:error_analysis}
\end{figure}

\subsection{Why does PoE with COT Fail?} \label{subsection:error_analysis}

To study the issues of PoE, we examine 50 rationales from CQA and ARC where PoE COT failed. We use our three largest LLMs, deemed the best at PoE. Our error types are: \textbf{1)~Misaligned Rationale:} Justifies a choice as correct instead of incorrect or fails to justify why its selection is incorrect; \textbf{2)~Reasoning Error:} Error in the reasoning chain, such as an inaccurate premise or conclusion; \textbf{3)~No Breakdown:} Restates choice without breakdown into reasoning steps; \textbf{4)~Hallucination:} References non-existent parts of the question; \textbf{5)~PoE N/A:} Question is unsuitable for PoE; and \textbf{6)~Dataset:} Dataset quality issue. Examples of rationales with each error type are in Appendix Table~\ref{appendix:error_table}.

In Figure~\ref{fig:error_analysis}, most errors stem from reasoning or misaligned rationales. The prevalence of these errors suggests that our PoE COT setting reveals the weaknesses of our tested LLMs to reason and follow in-context instructions, making our task a suitable testbed for these abilities. Notably, many misaligned rationales occur with negated questions (e.g. ``What would Grace \emph{not} do?''). LLMs struggle to reason under negation \cite{ravichander-etal-2022-condaqa} and since PoE is a negated reasoning technique, we may expect LLMs to underperform with this ``double negation.'' One solution to these issues is to fine-tune LLMs on PoE rationales. This may enhance PoE, but it would be interesting to see if this could also bolster overall reasoning capabilities.

\begin{figure}[t]
    \centering
    \includegraphics[width=\linewidth]{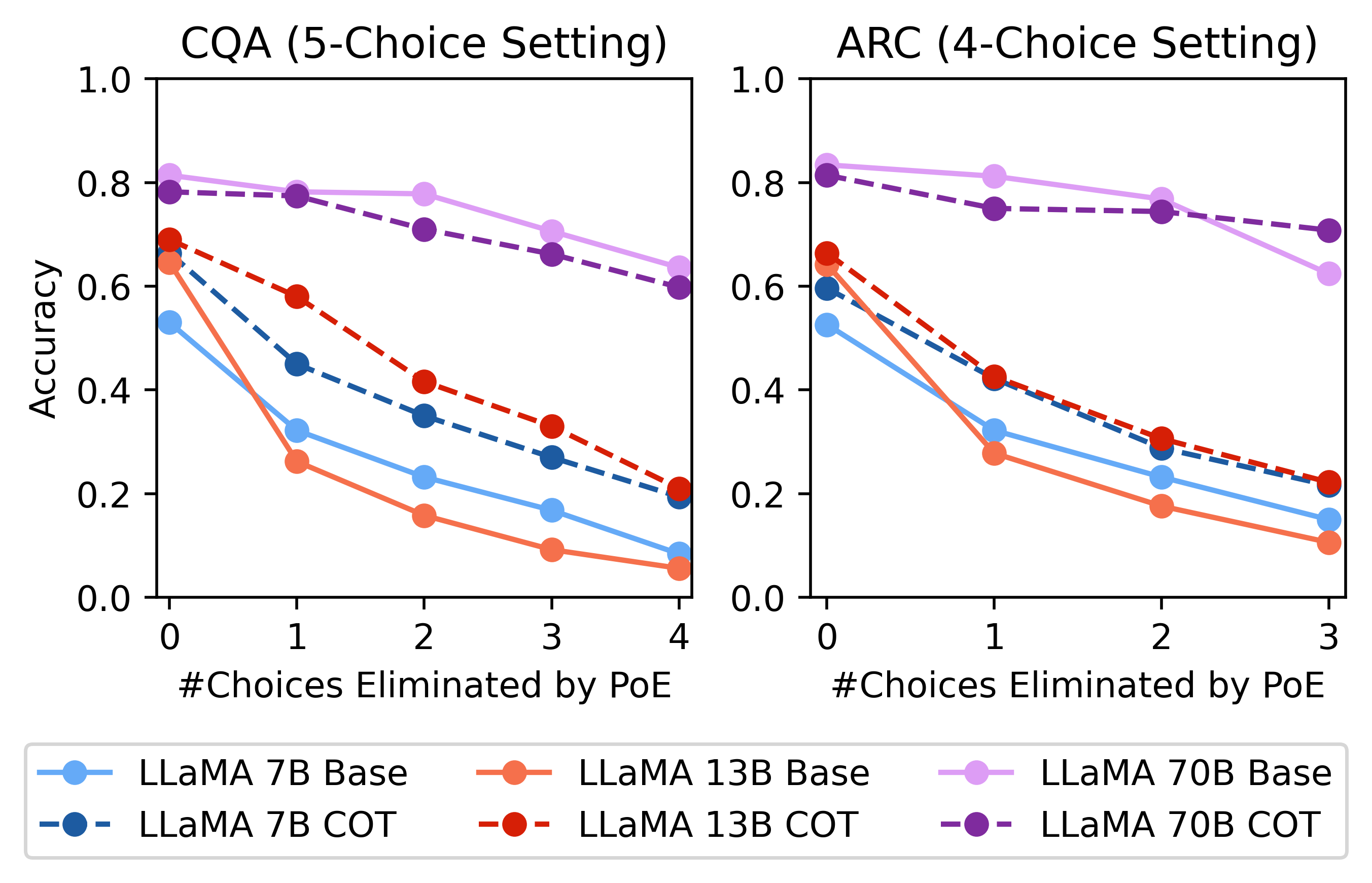}
    \caption{Accuracy of iterative PoE with each iteration.}
    \label{fig:iterative}
\end{figure}

\subsection{Is Iterative PoE Reliable?} \label{subsection:iterative_poe}

We speculated that if PoE fails with two choices, errors will propagate with full MC questions.
To confirm this, we run the setup in \cref{section:experimental_setup} on ARC/CQA with LLaMA, eliminating $i\in[0,n-1]$ choices and then selecting the correct answer.
The accuracy of this strategy decreases with each iteration both with PoE Base and COT (Figure~\ref{fig:iterative}). 
Thus, we suggest that future works improve the robustness of LLMs in 2-choice settings before trying iterative PoE, as iterative PoE currently leads to error propagation.




\section{Related Work}

\textbf{LLM Reasoning:} Several works explore if LLMs can reason with a chain-of-thought (COT) before giving a final answer \cite{wei2022chain, huang-chang-2023-towards}. Recent works on COT probe its faithfulness \cite{LYU2023FaithfulCR, Lanham2023MeasuringFI} and planning \cite{saparov2023language}. Similar to PoE with COT, prompt techniques like debate \cite{Michael2023DebateHS} and maieutic prompting \cite{jung2022maieutic} produce \emph{supporting} rationales for multiple options or let the model make its own decision, but we are the first to pinpoint if LLMs can reason why options are \emph{incorrect}, rather than correct. Appendix~\ref{section:related_work} discusses these works in-depth.




\noindent \textbf{Process of Elimination (PoE):} PoE has been studied in various settings. \citet{10.5555/3304222.3304364} train a neural model to perform PoE for reading comprehension. \citet{Zhang2023PREFERPE} and \citet{ma2023poe} use LMs to find the probability that candidates should be excluded for prompt ensembling and MCQA. \citet{tong2023eliminating} design a pipeline where an LLM proposes answers and removes incorrect proposals. While these works discard options to pick correct answers, we assess if generative LLMs can \emph{reason} toward incorrect options with COT. PoE with COT also has similarities with flipped classification label studies \cite{Wei2023LargerLM}, but we do not override the semantics of labels.

\section{Conclusion}

We study COT with process of elimination (PoE) and direct answer (DA) strategies in 2-choice commonsense and scientific reasoning datasets. We find that PoE underperforms DA in accuracy across all tested LLMs, but this gap narrows as model size scales. We also explore the logical inconsistencies between DA and PoE and categories of errors present in PoE rationales. Based on our results, we suggest several research directions: using PoE with COT as LLM reasoning/consistency benchmarks, combining DA and PoE with COT for calibration and refined rationales, and fine-tuning LLMs for PoE to enhance overall reasoning capabilities.

\section{Limitations}

LLMs are sensitive to prompts \cite{lu-etal-2022-fantastically}, so the accuracy of PoE COT could vary depending on the number of in-context examples and the human-written rationales in these prompts \cite{wei2022chain}. To mitigate this, we use an ample number (10) of in-context examples and write our prompts following the same format as \citet{wei2022chain}. Each rationale is written by one Ph.D. student and verified by a second Ph.D. student, both of whom work on LLM research, to ensure the rationales are high-quality. Rationales were also edited after running inference and identifying problems on a small validation set. Future works could explore rationale refinement or self-correction techniques \cite{zhang2023automatic, fu2023complexitybased, huang2023large} to have LLMs create their own prompts. 

Further, while PoE should ideally be able to give a rationale as to why one choice is correct without mentioning the other choice, this may be infeasible in certain scenarios. For example, given a mathematics question like ``What is 2+2*2?'' with the choices ``4'' and ``6'', the chain-of-thought leading the correct answer is clear (e.g. ``2 * 2 is 4 and 2 + 4 is 6, so the correct answer is 6''). However, it is very difficult to justify why ``4'' is incorrect without knowing that ``2+2*2=6''. To try to avoid these types of questions, we only look at scientific and commonsense reasoning questions, as we find it to be straightforward to explain why incorrect choices are incorrect. We also use these datasets to test PoE on questions with varying levels of objectivity (i.e. commonsense questions are more subjective than scientific questions). PoE is one of many test-taking strategies, so similar to students, future works could try teaching LLMs to choose the best strategies when answering multiple-choice questions, overcoming this limitation.

Lastly, due to resource constraints, we have not tested the ability of GPT-4 to perform PoE. In future studies related to PoE, it could be beneficial to evaluate GPT-4 to see if the LLM outperforms GPT-3.5 when executing PoE. If the gap between DA and PoE is smaller for GPT-4 than GPT-3.5, it would further support the idea that larger and more capable LLMs are more adept at performing PoE.

\section{Ethical Considerations}

Process of elimination is a strategy that aims to help LLMs eliminate incorrect options, with the goal of promoting interpretability and self-consistency, as well as enhancing downstream applications. However, our current findings suggest that PoE with COT may be an unreliable technique. Thus, we advise researchers and LLM practitioners to take caution before deploying PoE in any application.

Further, as with all reasoning techniques, researchers should ensure that PoE rationales are not based on stereotypes or biases. This is especially important in some of the downstream applications we mention in this work, like medical diagnoses of exclusion, where the backgrounds of certain individuals could introduce bias into the LLM's reasoning. We hope that future research endeavors will jointly attempt to address these biases and improve the robustness of LLM reasoning, fostering more equitable decision-making frameworks.

\section{Acknowledgements}

We would like to thank members of the CLIP lab at the University of Maryland and external collaborators for their feedback and discussions of this work, including Dayeon (Zoey) Ki, Yu (Hope) Hou, Yoo Yeon Sung, Shi Feng, and Jordan Boyd-Graber.
We also thank the anonymous reviewers for their feedback.
This material is based upon work supported by the
National Science Foundation Graduate Research Fellowship Program under Grant No. DGE 2236417. Any opinions, findings, and conclusions or recommendations expressed in this material
are those of the author(s) and do not necessarily reflect the views of the National Science Foundation.

\bibliography{custom}
\bibliographystyle{acl_natbib}

\appendix \label{sec:appendix}

\clearpage

\section{Experimental Setup Details}

\subsection{Dataset Statistics} \label{appendidx:dataset}

In Table~\ref{appendix:dataset_stats}, we display summary statistics for the datasets used in our experiments. All datasets are publicly available and free to use on HuggingFace.

\subsection{Prompt Details} \label{appendidx:prompts}

\begin{figure*}
    \centering
    \fbox{\includegraphics[width=0.75\linewidth]{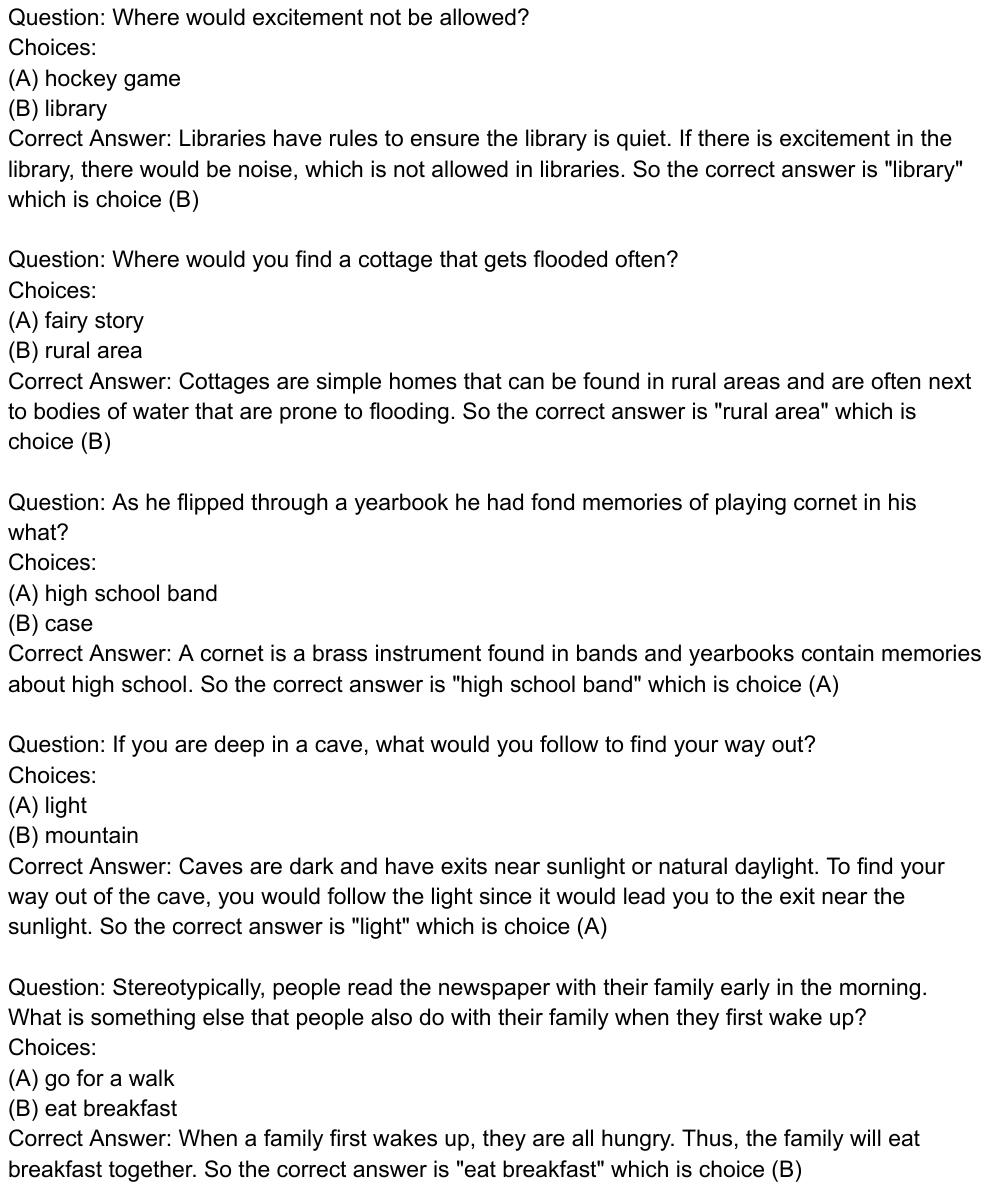}}
    \caption{\label{appendix:cqa_da} Five in-context learning examples for DA COT on Commonsense QA}
\end{figure*}

\begin{figure*}
    \centering
    \fbox{\includegraphics[width=0.75\linewidth]{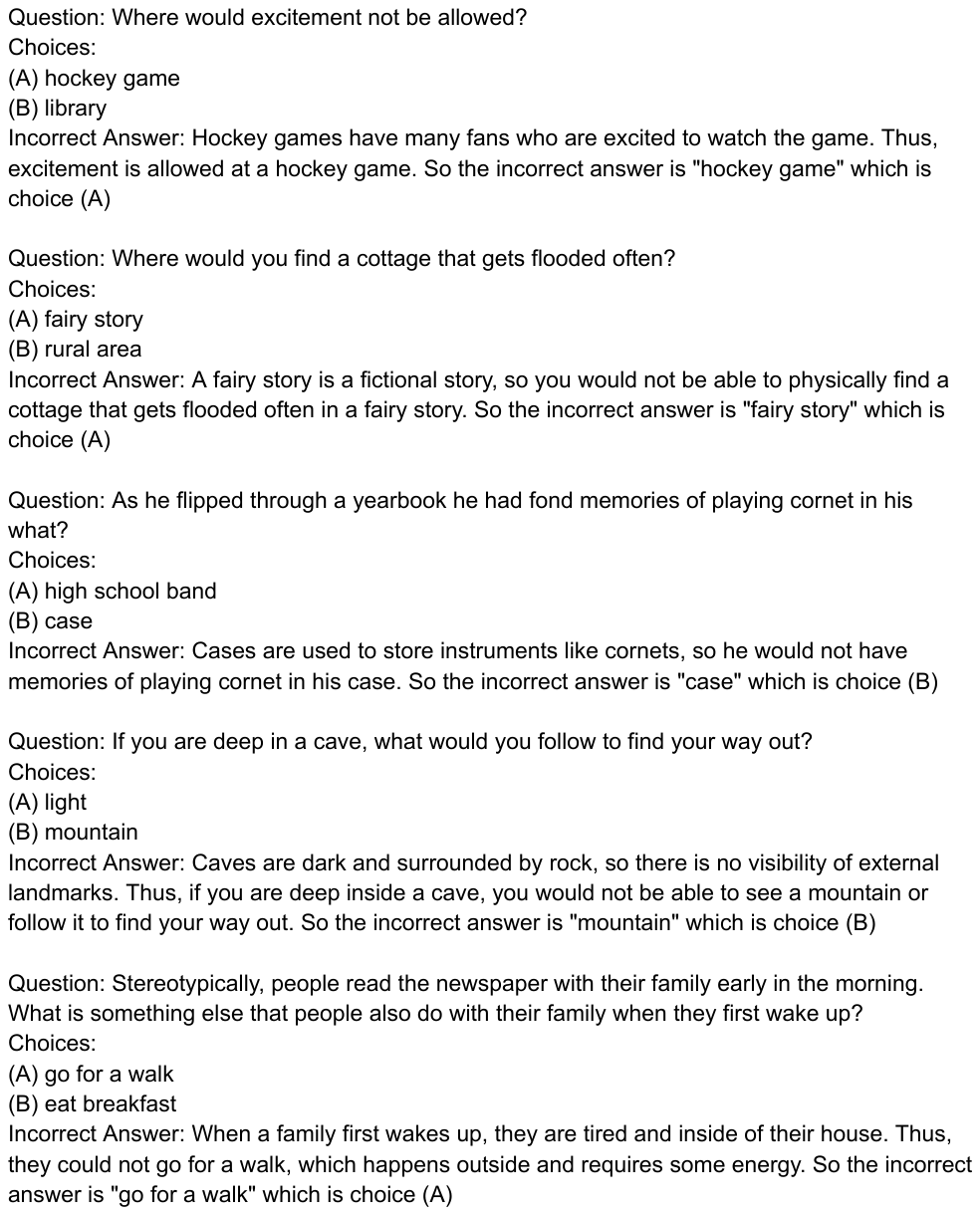}}
    \caption{\label{appendix:cqa_PoE} Five in-context learning examples for PoE COT on Commonsense QA}
\end{figure*}

\begin{figure*}
    \centering
    \fbox{\includegraphics[width=0.75\linewidth]{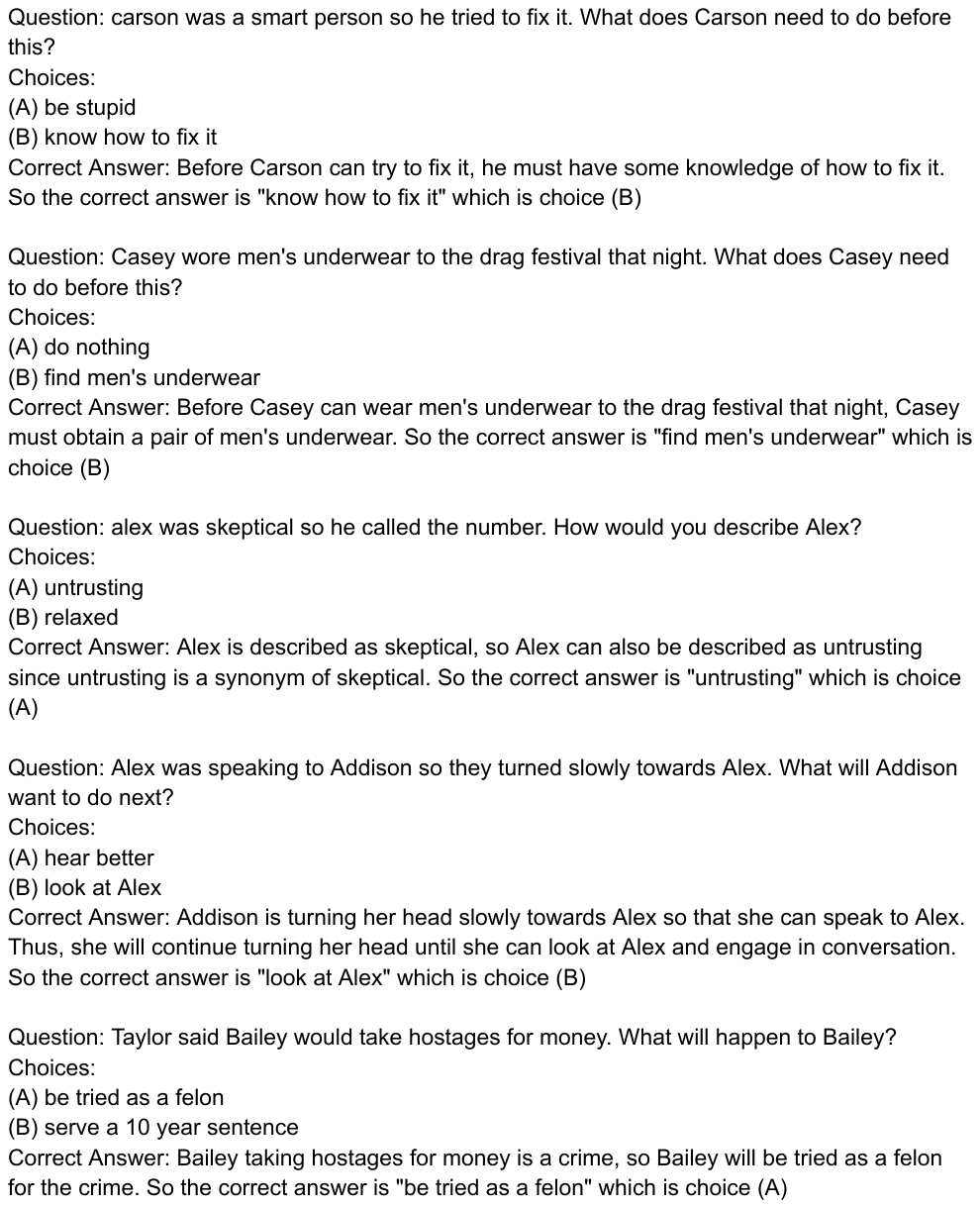}}
    \caption{\label{appendix:siqa_da} Five in-context learning examples for DA COT on Social IQa}
\end{figure*}

\begin{figure*}
    \centering
    \fbox{\includegraphics[width=0.75\linewidth]{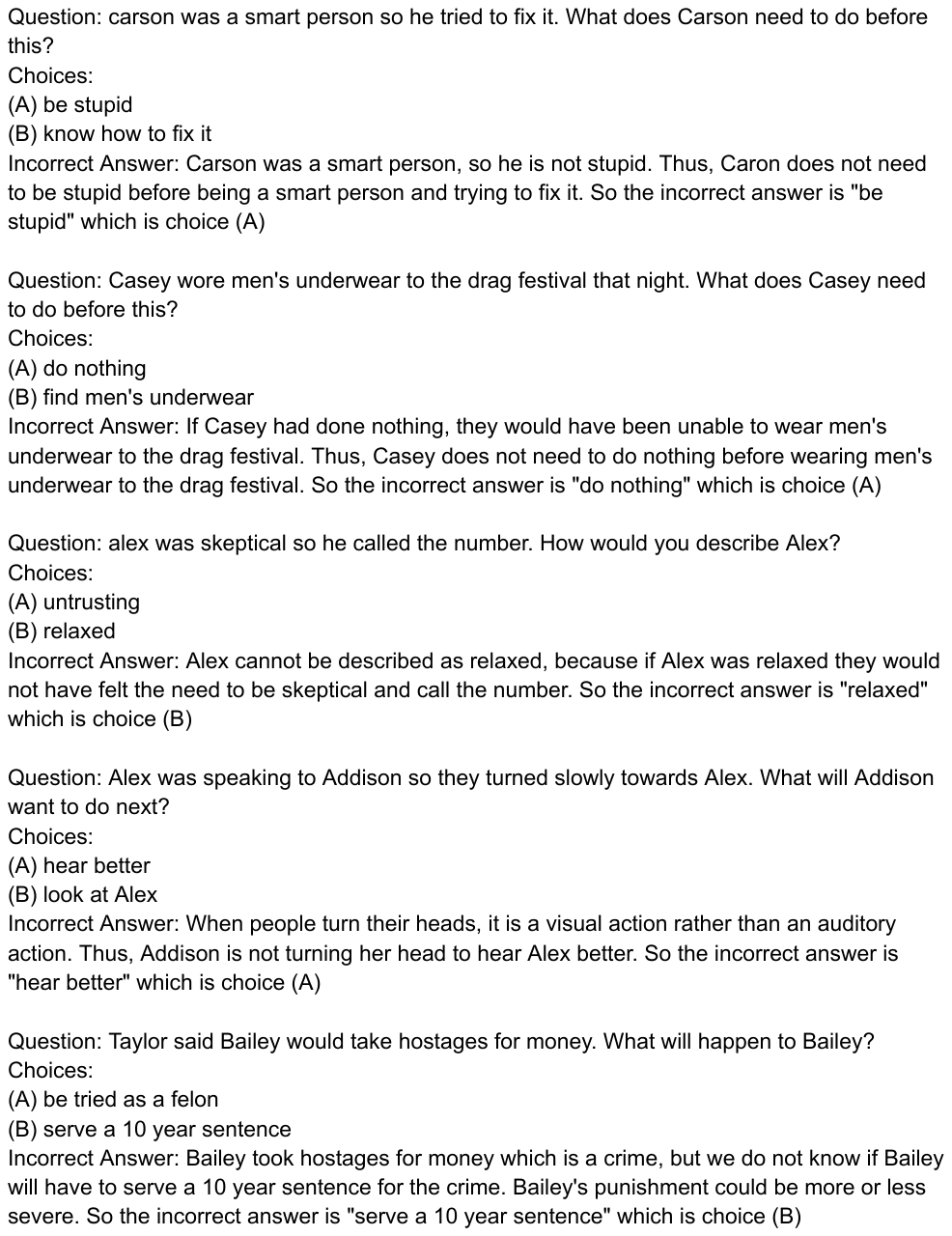}}
    \caption{\label{appendix:siqa_PoE} Five in-context learning examples for PoE COT on Social IQa}
\end{figure*}

\begin{figure*}
    \centering
    \fbox{\includegraphics[width=0.75\linewidth]{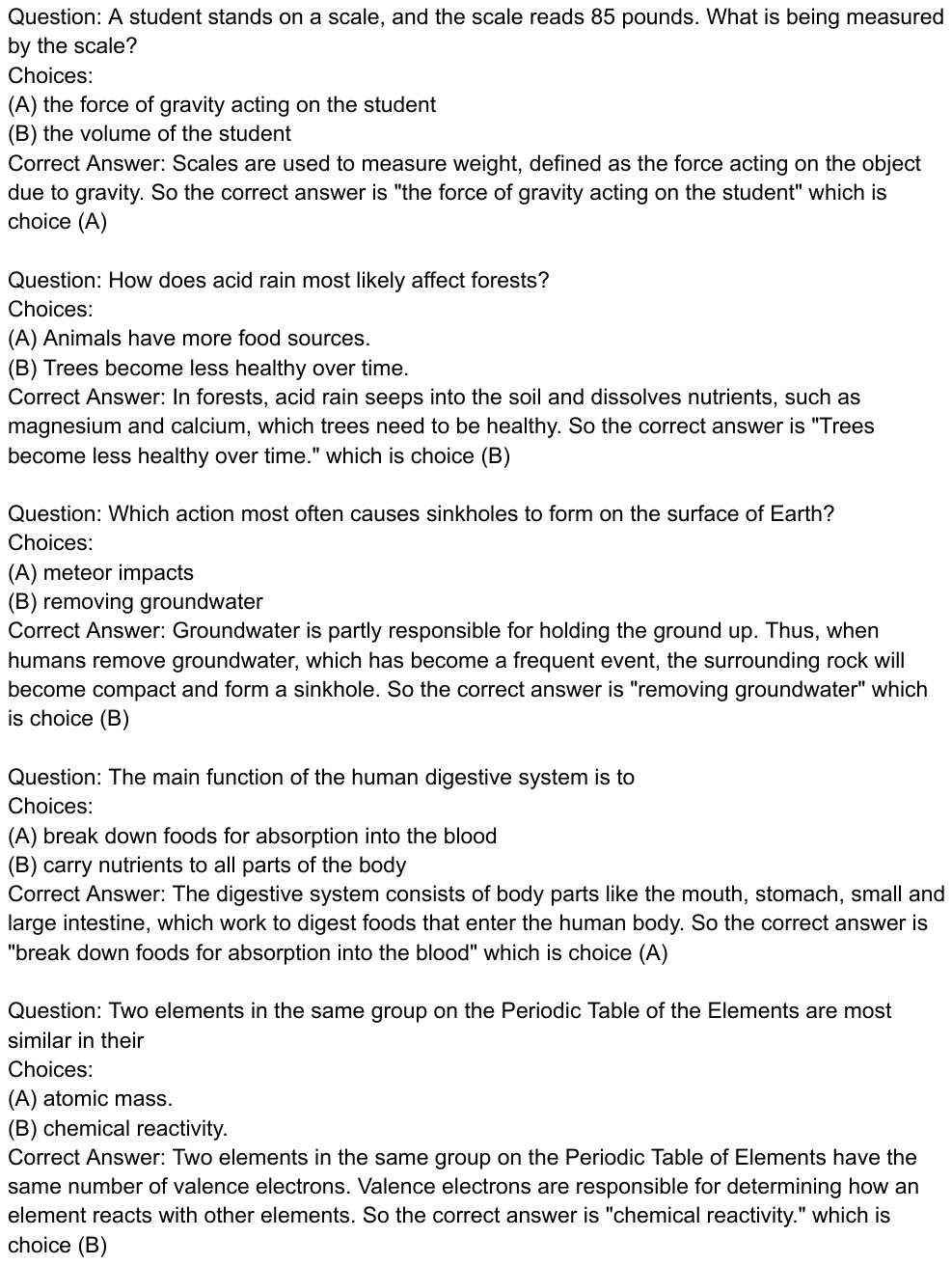}}
    \caption{\label{appendix:arc_da} Five in-context learning examples for DA COT on ARC}
\end{figure*}

\begin{figure*}
    \centering
    \fbox{\includegraphics[width=0.75\linewidth]{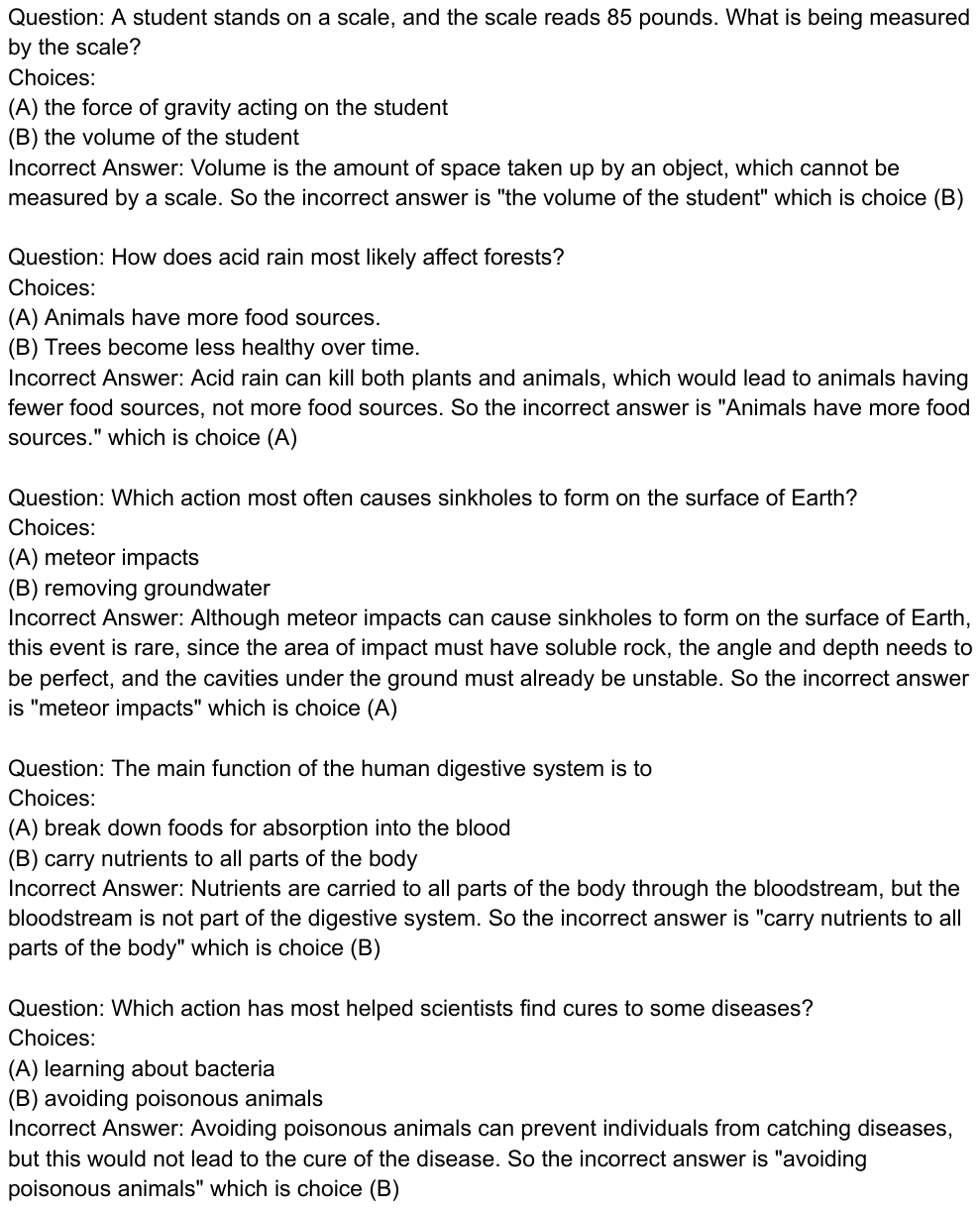}}
    \caption{\label{appendix:arc_PoE} Five in-context learning examples for PoE COT on ARC}
\end{figure*}

\begin{figure*}
    \centering
    \fbox{\includegraphics[width=0.75\linewidth]{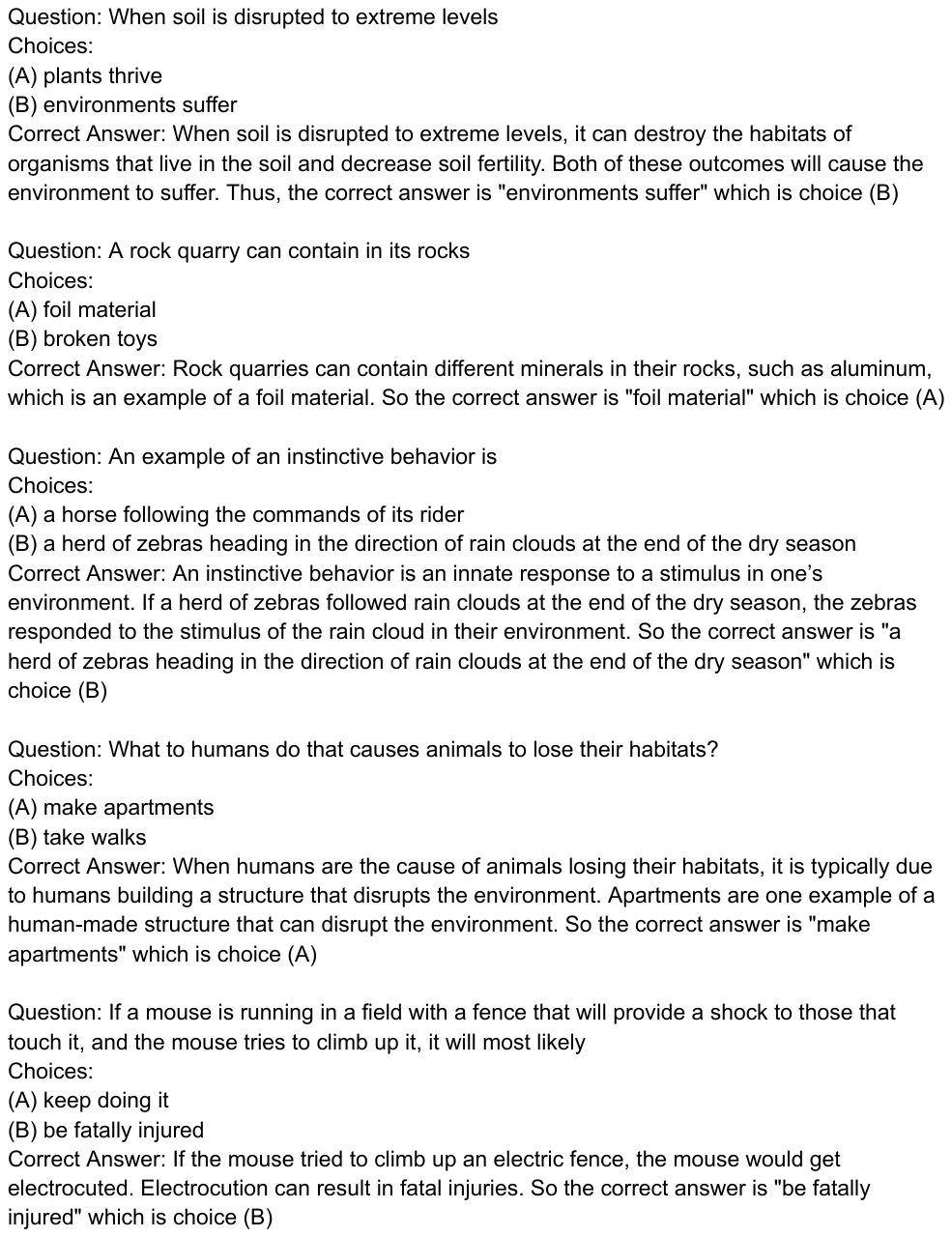}}
    \caption{\label{appendix:obqa_da} Five in-context learning examples for DA COT on OpenBookQA}
\end{figure*}

\begin{figure*}
    \centering
    \fbox{\includegraphics[width=0.75\linewidth]{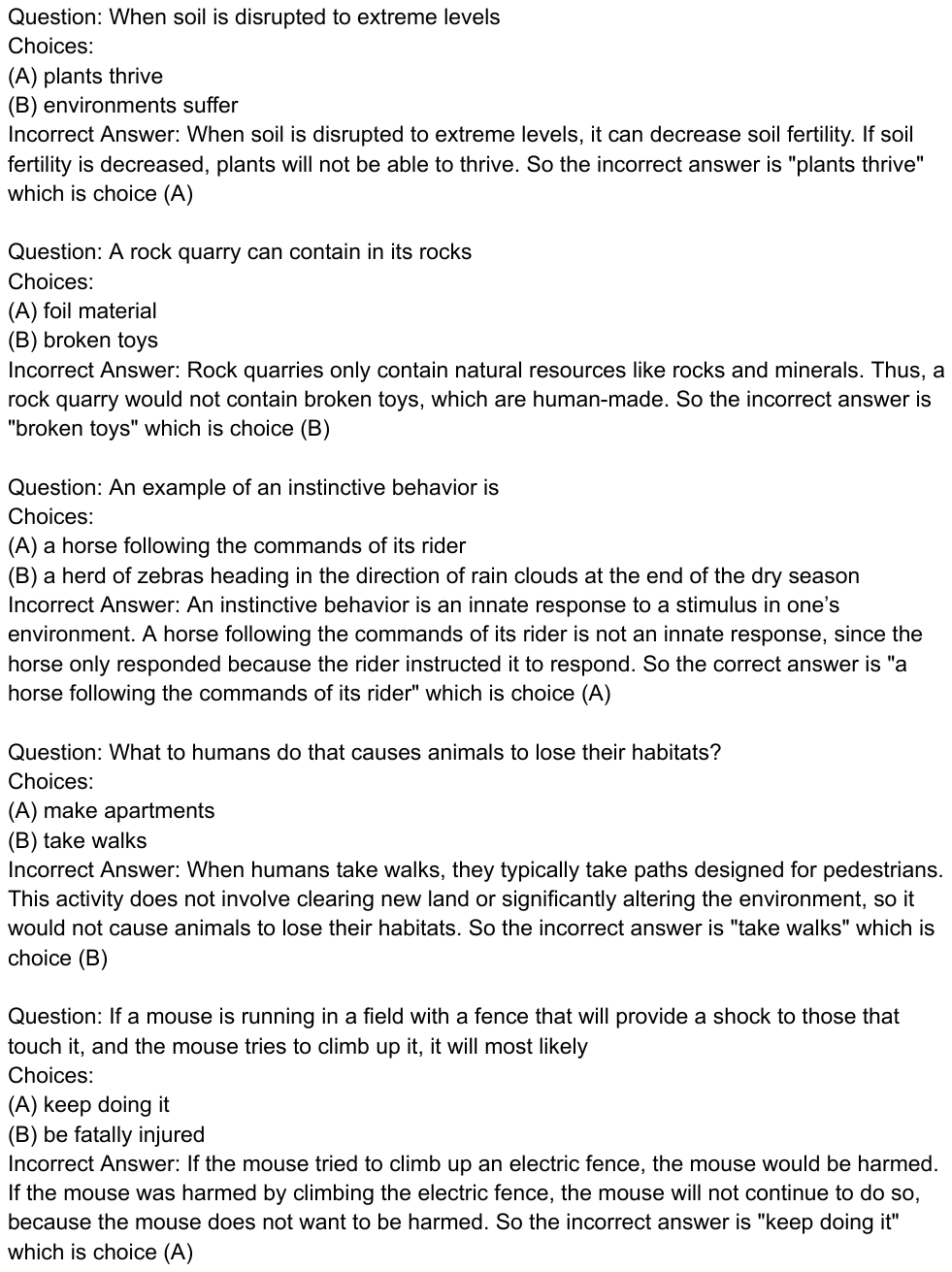}}
    \caption{\label{appendix:obqa_PoE} Five in-context learning examples for PoE COT on OpenBookQA}
\end{figure*}

Examples of the prompts used for DA COT and PoE COT on each dataset can be found in Figures~\ref{appendix:cqa_da}, \ref{appendix:cqa_PoE}, \ref{appendix:siqa_da}, \ref{appendix:siqa_PoE}, \ref{appendix:arc_da}, \ref{appendix:arc_PoE}, \ref{appendix:obqa_da}, and \ref{appendix:obqa_PoE}. The Base versions of these prompts follow the same format, except only the answer choice follows the ``Correct Answer:'' and ``Incorrect Answer:'' labels. For example, if the question was ``Where is Chicago?'' with choices ``Illinois'' and ``the Moon,'' the DA Base prompt would be: \\

\noindent \texttt{Question: Where is Chicago?\\Choices:\\(A) Illinois\\(B) The Moon\\Correct Answer: (A)} \\ 

\noindent and the PoE Base prompt would be: \\

\noindent \texttt{Question: Where is Chicago?\\Choices:\\(A) Illinois\\(B) The Moon\\Incorrect Answer: (B)} \\ 

Before the in-context learning examples, we prepend the instruction ``\emph{Your goal is to identify the correct answer to the multiple choice question}'' to the DA prompts, and the instruction ``\emph{Your goal is to identify the incorrect answer to the multiple choice question}'' to the PoE prompts. For the 10 total in-context examples, five of the examples select choice (A) and five of the examples select choice (B), which are shuffled. This ensures that our few-shot prompts are as strong as possible. We release the full prompts along with our code.

\subsection{Model Implementation}

The 7B open-source LLMs were run on a single NVIDIA RTX A6000 GPU. The other open-source LLMs were run on 8 NVIDIA RTX A5000 GPUs. GPT-3.5 was run using CPU only. We allocated 24 hours for a single LLM to run all four strategies (DA Base, DA COT, PoE Base, PoE COT) on all four datasets. The experiments from Figure~\ref{fig:benchmark} are from a single run, while the experiments from Table~\ref{table:consistency} require up to two runs.

\section{Detailed Results}

\subsection{Detailed Quantitative Evaluation} \label{appendix:numerical_eval}

In Tables~\ref{table:CQA_eval}, \ref{table:SIQA_eval}, \ref{table:ARC_eval}, and \ref{table:openbook_eval}, we provide the numerical results from Figure~\ref{fig:benchmark}. We also calculate the difference in accuracies of the DA and POE strategies, along with their associated $p$-values. We find that a majority of the differences in accuracy between DA and PoE are statistically significant.

\subsection{Full Consistency Experiments} \label{appendidx:consistency}

We report the consistency results from \cref{subsection:consistent} for all LLMs and datasets. In addition to the self-consistency of DA, we include the self-consistency of PoE to ensure that the lack of consistency between DA and PoE is not due to higher sampling variance for PoE. In Tables~\ref{table:consistency_detailed_llama-2_7b}, \ref{table:consistency_detailed_llama-2_13b}, and \ref{table:consistency_detailed_llama-2_70b}, we report the results for LLaMA-2. In Tables~\ref{table:consistency_detailed_falcon_7b}, \ref{table:consistency_detailed_falcon_40b}, and \ref{table:consistency_detailed_falcon_180b}, we report the results for Falcon. In Table~\ref{table:consistency_detailed_gpt-35}, we report the results for GPT-3.5. 

We find that the agreement of DA and PoE is the lowest across all datasets and LLMs, except for LLaMA-2 70B on Commonsense QA. We also note that PoE self-consistency is typically lower than DA self-consistency, meaning that PoE has a higher variance which further points to its unreliability.

\subsection{Qualitative Results}

\begin{figure*}[t]
    \centering
    \includegraphics[width=\linewidth]{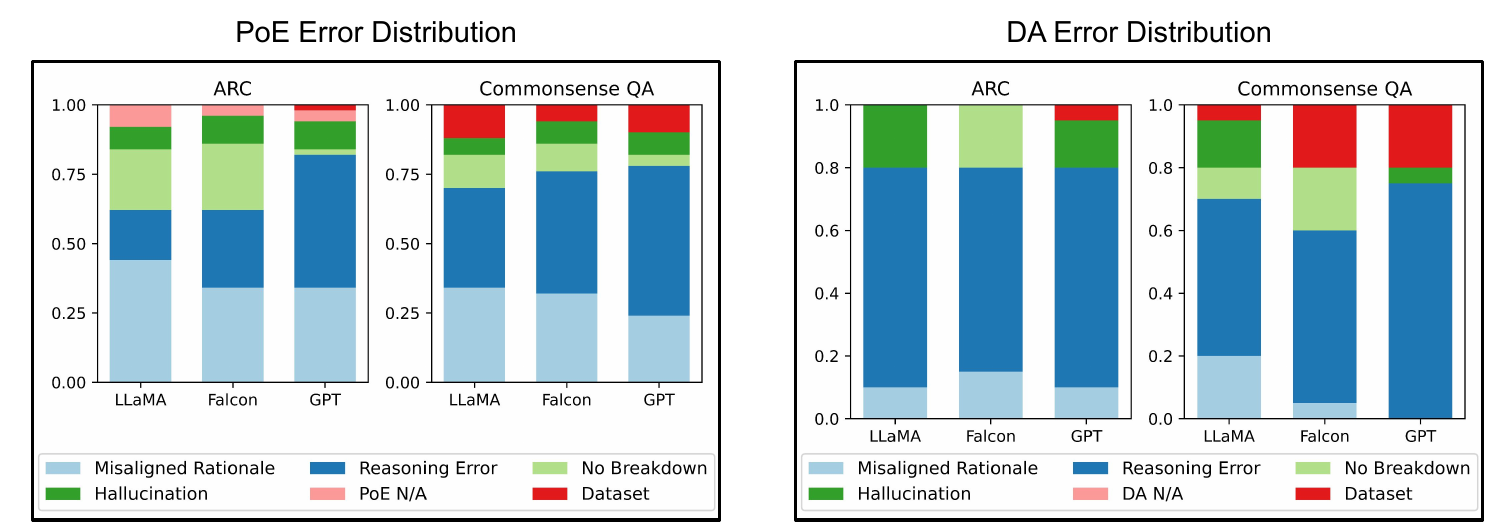}
    \caption{Error distribution of PoE COT and DA COT on ARC and Commonsense QA}
    \label{appendix:error_analysis_full}
\end{figure*}

Examples of PoE rationales that fall into the categories of our error analysis can be found in Table~\ref{appendix:error_table}. Examples of sound rationales generated by PoE COT can be found in Table~\ref{appendix:success_table}. In Figure~\ref{appendix:error_analysis_full}, we compare the distribution of errors of PoE COT and DA COT. For DA COT, we use the same error categories defined in \cref{subsection:error_analysis}, except ``PoE N/A'' is replaced with ``DA N/A''. Further, we only evaluate 20 DA COT rationales, since there were fewer instances of errors from DA COT to choose from.

The majority of the DA COT errors were reasoning errors (and not misaligned rationales), suggesting that the LLMs better understand the DA task compared to PoE. This may further indicate that LLMs have a training bias toward identifying correct answers. Further, we find that many of the misaligned rationale errors that occur with DA COT are also due to negated questions (e.g. ``What would Grace not do?''), reinforcing the difficulty of LLMs to properly address questions with negation.

\subsection{Iterative PoE Details} \label{appendix:iterative_poe}

In this section, we provide more details on our setup for the iterative PoE experiment in \cref{subsection:iterative_poe}. If a question has $n$ choices, we independently prompt the LLM using PoE Base and PoE COT to eliminate an answer $n-1$ times, so the model is unaware of its previous decisions. At each step when there are $i \in [2, n]$ choices, we also run the DA Base and DA COT prompts, corresponding to every possible use case of PoE (i.e. \emph{eliminate $i$ choices and then choose the correct answer}). PoE Base only is used with DA Base, and PoE COT is only used with DA COT.

In Tables \ref{appendix:cqa_iterative} and \ref{appendix:arc_iterative}, we display the results for iterative PoE on Commonsense QA and ARC, respectively. Overall, we find that the raw accuracy decreases significantly as more choices get eliminated, confirming our intuition that error propagation is an issue in this setting. Although PoE has higher accuracy than DA when the number of choices remaining is higher, this finding is not that significant when you consider the probability of performing each strategy accurately by chance. For example, when there are 5 choices, there is a probability of $\frac{4}{5} = 0.8$ of guessing an incorrect answer correctly, while there is only a probability of $\frac{1}{5} = 0.2$ of guessing the correct answer correctly. If future research eventually leads to LLMs that can perform the DA and PoE strategies at similar abilities, it would be interesting to revisit the reliability and effectiveness of iterative PoE.

\subsection{Why not compare with score-based PoE techniques?}

As described in the related work, previous works have explored PoE as a means to discard low-confidence options when selecting correct answers \cite{Zhang2023PREFERPE, ma2023poe}. These works execute PoE by first prompting an LM to choose the correct answer, and obtain a distribution of token probabilities over all of the options. The answers with token probabilities below a certain threshold are discarded. For example, \citet{ma2023poe} discard answers with probabilities that are lower than the average probability of all answers.

This approach, however, is not very useful in the two-choice setting that we study, as the model will always be consistent. To illustrate, in a 2-choice setting, the LM will ascribe choice ``A'' probability $a$ and choice ``B'' probability $b$ (where $a+b=1$). Assuming $a > b$ without loss of generality, a direct answer strategy in this context would always select choice ``A'' (since $a>b$), and PoE would always eliminate choice ``B'' (since $b < \frac{1}{2} (a + b)$). 

Thus, comparing with these score-based PoE techniques does not give us any useful signals into the reasoning and decision-making capabilities of LLMs, as the model will always be logically consistent. In the PoE and DA setup that we describe in this work, models have the potential to be logically inconsistent, providing deeper insights into the true decision-making capabilities of our tested LLMs. Further, the aspect of COT allows us to interpret the rationales from models (\cref{subsection:error_analysis}), which cannot be accomplished with score-based PoE techniques. 

\section{Related Work} \label{section:related_work}

There exist a set of techniques, such as Maieutic Prompting (MP) \cite{jung2022maieutic}, that also employ LLMs to generate rationales for multiple options in multiple-choice question answering. Below, we describe the novelty of PoE with COT and how our task is different from the setup of MP.

In MP, the LLM is asked to generate unique rationales of why a True/False question (e.g. \emph{War cannot have a tie?}) is true and why it is false. In other words, MP forces the model to generate \textbf{supporting} rationales for both a correct answer and an incorrect answer. This is subtly but critically distinct from our setup. In PoE with CoT, the model is given a (non-True/False) question with two answer choices, and is asked to select the incorrect choice and provide a rationale for why it is incorrect. In short, MP forces a model to explain why an incorrect (T/F) answer is \emph{correct}, while we ask the model to select an incorrect (non-T/F) answer and explain why it is \emph{incorrect}.

For example, given the T/F question “Is 1 greater than 2,” MP would ask a model to explain (impossibly) why it is true that 1 is greater than 2. Conversely, given the non-T/F question “Where do you put your grapes before checking out,” PoE would ask a model to reason why you don’t put grapes in your mouth before paying at a grocery store (per Figure~\ref{fig:intro}). These setups are inherently different.

Other techniques, such as debate \cite{Michael2023DebateHS}, also fall into this same category as MP. Further, in debate, is made even more explicit in debate that LLMs may be unreliable and inaccurately arguing for an incorrect option. In contrast, PoE with COT tasks LLMs with generating \emph{accurate} rationales as to why an answer is incorrect.


\begin{table*}[t]
\small
\centering
\setlength{\tabcolsep}{4pt}
\begin{tabular}{@{}c|c|c|c|c|c|c@{}}
\toprule
\multicolumn{1}{c|}{\textbf{Dataset}} & \textbf{Category} & \textbf{\# Questions} & \textbf{\# Choices} & \textbf{\specialcell{Proportion where\\Gold Answer\\is Choice (A)}} & \textbf{\specialcell{Avg Question\\Length}} & \textbf{\specialcell{Avg Choice\\Length}} \\ \midrule
Commonsense QA & Commonsense Reasoning & 500 & 2 & 0.514 & 15.60 & 2.06 \\
Social IQa & Commonsense Reasoning & 500 & 2 & 0.490 & 24.21 & 4.09 \\
ARC & Scientific Reasoning & 500 & 2 & 0.486 & 25.78 & 6.31 \\
OpenBook QA & Scientific Reasoning & 500 & 2 & 0.495 & 11.72 & 3.72 \\ \bottomrule
\end{tabular}
\caption{Summary statistics for the datasets used in our experiments. Average length is calculated with the GPT-4 tokenizer, implemented through the tiktoken library.}
\label{appendix:dataset_stats}
\end{table*}

\begin{table*}[t]
\centering
\small
\begin{tabular}{@{}l|cccc|cccc@{}}
\toprule
\multicolumn{1}{c|}{\multirow{2}{*}{\textbf{Model}}} & \multicolumn{4}{c|}{\textbf{Base}} & \multicolumn{4}{c}{\textbf{COT}} \\
\multicolumn{1}{c|}{} & DA Base & PoE Base & Base Diff & Base $p$-val & DA COT & PoE COT & COT Diff & COT $p$-val  \\ \midrule
LLaMA 7B & 0.816 & 0.320 & 0.496 & \underline{0.000} & 0.862 & 0.458 & 0.404 & \underline{0.000} \\
LLaMA 13B & 0.858 & 0.266 & 0.592 & \underline{0.000} & 0.868 & 0.421 & 0.447 & \underline{0.000} \\
LLaMA 70B & 0.930 & 0.846 & 0.084 & \underline{0.000} & 0.916 & 0.808 & 0.108 & \underline{0.000} \\ \midrule
Falcon 7B & 0.516 & 0.472 & 0.044 & 0.164 & 0.690 & 0.466 & 0.224 & \underline{0.000} \\
Falcon 40B & 0.892 & 0.740 & 0.152 & \underline{0.000} & 0.876 & 0.532 & 0.344 & \underline{0.000} \\
Falcon 180B & 0.924 & 0.912 & 0.012 & 0.490 & 0.926 & 0.786 & 0.140 & \underline{0.000} \\ \midrule
GPT 3.5 & 0.890 & 0.778 & 0.112 & \underline{0.000} & 0.937 & 0.853 & 0.084 & \underline{0.000} \\ \bottomrule
\end{tabular}
\caption{\label{table:CQA_eval}Evaluation of Direct Answer (DA) and Process of Elimination (PoE) strategies on Commonsense QA, with and without chain-of-thought reasoning. Diff denotes the difference between the accuracies of the two strategies. $p$-val corresponds to a difference in means t-test. \underline{Underlined} values denote $p$-val $\leq$ 0.05}
\end{table*}

\begin{table*}[t]
\centering
\small
\begin{tabular}{@{}l|cccc|cccc@{}}
\toprule
\multicolumn{1}{c|}{\multirow{2}{*}{\textbf{Model}}} & \multicolumn{4}{c|}{\textbf{Base}} & \multicolumn{4}{c}{\textbf{COT}} \\
\multicolumn{1}{c|}{} & DA Base & PoE Base & Base Diff & Base $p$-val & DA COT & PoE COT & COT Diff & COT $p$-val  \\ \midrule
LLaMA 7B & 0.762 & 0.440 & 0.322 & \underline{0.000} & 0.758 & 0.548 & 0.210 & \underline{0.000} \\
LLaMA 13B & 0.772 & 0.408 & 0.364 & \underline{0.000} & 0.806 & 0.578 & 0.228 & \underline{0.000} \\
LLaMA 70B & 0.876 & 0.814 & 0.062 & \underline{0.007} & 0.853 & 0.796 & 0.057 & \underline{0.019} \\ \midrule
Falcon 7B & 0.518 & 0.500 & 0.018 & 0.570 & 0.651 & 0.483 & 0.168 & \underline{0.000} \\
Falcon 40B & 0.808 & 0.718 & 0.090 & \underline{0.001} & 0.820 & 0.629 & 0.190 & \underline{0.000} \\
Falcon 180B & 0.858 & 0.802 & 0.056 & \underline{0.018} & 0.842 & 0.717 & 0.124 & \underline{0.000} \\ \midrule
GPT 3.5 & 0.838 & 0.820 & 0.018 & 0.450 & 0.865 & 0.748 & 0.117 & \underline{0.000} \\ \bottomrule
\end{tabular}
\caption{\label{table:SIQA_eval}Evaluation of Direct Answer (DA) and Process of Elimination (PoE) strategies on Social IQa, with and without chain-of-thought reasoning. Diff denotes the difference between the accuracies of the two strategies. $p$-val corresponds to a difference in means t-test. \underline{Underlined} values denote $p$-val $\leq$ 0.05}
\end{table*}

\begin{table*}[t]
\centering
\small
\begin{tabular}{@{}l|cccc|cccc@{}}
\toprule
\multicolumn{1}{c|}{\multirow{2}{*}{\textbf{Model}}} & \multicolumn{4}{c|}{\textbf{Base}} & \multicolumn{4}{c}{\textbf{COT}} \\
\multicolumn{1}{c|}{} & DA Base & PoE Base & Base Diff & Base $p$-val & DA COT & PoE COT & COT Diff & COT $p$-val  \\ \midrule
LLaMA 7B & 0.770 & 0.478 & 0.292 & \underline{0.000} & 0.787 & 0.494 & 0.293 & \underline{0.000} \\
LLaMA 13B & 0.816 & 0.266 & 0.550 & \underline{0.000} & 0.832 & 0.579 & 0.253 & \underline{0.000} \\
LLaMA 70B & 0.930 & 0.744 & 0.186 & \underline{0.000} & 0.938 & 0.878 & 0.060 & \underline{0.001} \\ \midrule
Falcon 7B & 0.544 & 0.470 & 0.074 & \underline{0.019} & 0.546 & 0.422 & 0.124 & \underline{0.000} \\
Falcon 40B & 0.836 & 0.656 & 0.180 & \underline{0.000} & 0.814 & 0.546 & 0.267 & \underline{0.000} \\
Falcon 180B & 0.938 & 0.874 & 0.064 & \underline{0.001} & 0.912 & 0.822 & 0.089 & \underline{0.000} \\ \midrule
GPT 3.5 & 0.914 & 0.856 & 0.058 & \underline{0.004} & 0.948 & 0.855 & 0.092 & \underline{0.000} \\ \bottomrule
\end{tabular}
\caption{\label{table:ARC_eval}Evaluation of Direct Answer (DA) and Process of Elimination (PoE) strategies on ARC, with and without chain-of-thought reasoning. Diff denotes the difference between the accuracies of the two strategies. $p$-val corresponds to a difference in means t-test. \underline{Underlined} values denote $p$-val $\leq$ 0.05}
\end{table*}

\begin{table*}[t]
\centering
\small
\begin{tabular}{@{}l|cccc|cccc@{}}
\toprule
\multicolumn{1}{c|}{\multirow{2}{*}{\textbf{Model}}} & \multicolumn{4}{c|}{\textbf{Base}} & \multicolumn{4}{c}{\textbf{COT}} \\
\multicolumn{1}{c|}{} & DA Base & PoE Base & Base Diff & Base $p$-val & DA COT & PoE COT & COT Diff & COT $p$-val  \\ \midrule
LLaMA 7B & 0.790 & 0.399 & 0.391 & \underline{0.000} & 0.851 & 0.594 & 0.257 & \underline{0.000} \\
LLaMA 13B & 0.818 & 0.277 & 0.541 & \underline{0.000} & 0.857 & 0.529 & 0.328 & \underline{0.000} \\
LLaMA 70B & 0.938 & 0.677 & 0.261 & \underline{0.000} & 0.924 & 0.824 & 0.100 & \underline{0.000} \\ \midrule
Falcon 7B & 0.523 & 0.483 & 0.040 & 0.206 & 0.632 & 0.453 & 0.179 & \underline{0.000} \\
Falcon 40B & 0.848 & 0.747 & 0.100 & \underline{0.000} & 0.869 & 0.685 & 0.185 & \underline{0.000} \\
Falcon 180B & 0.924 & 0.888 & 0.036 & 0.051 & 0.932 & 0.812 & 0.120 & \underline{0.000} \\ \midrule
GPT 3.5 & 0.892 & 0.745 & 0.146 & \underline{0.000} & 0.911 & 0.853 & 0.058 & \underline{0.004} \\ \bottomrule
\end{tabular}
\caption{\label{table:openbook_eval}Evaluation of Direct Answer (DA) and Process of Elimination (PoE) strategies on OpenBookQA, with and without chain-of-thought reasoning. Diff denotes the difference between the accuracies of the two strategies. $p$-val corresponds to a difference in means t-test. \underline{Underlined} values denote $p$-val $\leq$ 0.05}
\end{table*}

\begin{table*}[t]
\centering
\small
\begin{tabular}{@{}ccccccc@{}}
 & \multicolumn{6}{c}{\emph{LLaMA-2 7B}} \\ \midrule
\multicolumn{1}{c|}{\multirow{2}{*}{Dataset}} & \multicolumn{3}{c|}{\textbf{Base}} & \multicolumn{3}{c}{\textbf{COT}}  \\
\multicolumn{1}{c|}{}  & DA Self & PoE Self & \multicolumn{1}{c|}{DA vs. PoE}  & DA Self & PoE Self & DA vs. PoE  \\ \midrule
\multicolumn{1}{c|}{Commonsense QA} & 90.6 & 78.8 & \multicolumn{1}{c|}{\underline{37.4}} & 88.2 & 68.1 & \underline{47.9} \\ 
\multicolumn{1}{c|}{Social IQa} & 90.0 & 74.6 & \multicolumn{1}{c|}{\underline{44.0}} & 83.2 & 66.2 & \underline{52.8} \\ \midrule
\multicolumn{1}{c|}{ARC} & 89.0 & 74.1 & \multicolumn{1}{c|}{\underline{51.0}} & 84.6 & 71.2 & \underline{45.3} \\ 
\multicolumn{1}{c|}{OpenBookQA} & 86.3 & 77.9 & \multicolumn{1}{c|}{\underline{35.1}} & 84.7 & 70.5 & \underline{52.7} \\ \bottomrule
\end{tabular}

\caption{\label{table:consistency_detailed_llama-2_7b}Self-Consistency of the DA and PoE strategies versus the consistency of DA and PoE for LLaMA-2 7B. Least consistent methods are \underline{underlined}.}
\end{table*}

\begin{table*}[t]
\centering
\small
\begin{tabular}{@{}ccccccc@{}}
 & \multicolumn{6}{c}{\emph{LLaMA-2 13B}} \\ \midrule
\multicolumn{1}{c|}{\multirow{2}{*}{Dataset}} & \multicolumn{3}{c|}{\textbf{Base}} & \multicolumn{3}{c}{\textbf{COT}}  \\
\multicolumn{1}{c|}{}  & DA Self & PoE Self & \multicolumn{1}{c|}{DA vs. PoE}  & DA Self & PoE Self & DA vs. PoE  \\ \midrule
\multicolumn{1}{c|}{Commonsense QA} & 92.6 & 74.4 & \multicolumn{1}{c|}{\underline{25.3}} & 91.2 & 69.1 & \underline{42.3} \\ 
\multicolumn{1}{c|}{Social IQa} & 92.7 & 64.6 & \multicolumn{1}{c|}{\underline{42.4}} & 82.8 & 71.0 & \underline{55.3} \\ \midrule
\multicolumn{1}{c|}{ARC} & 92.8 & 75.9 & \multicolumn{1}{c|}{\underline{21.7}} & 87.5 & 71.4 & \underline{55.4} \\ 
\multicolumn{1}{c|}{OpenBookQA} & 86.5 & 76.7 & \multicolumn{1}{c|}{\underline{23.7}} & 87.9 & 69.8 & \underline{50.2} \\ \bottomrule
\end{tabular}

\caption{\label{table:consistency_detailed_llama-2_13b}Self-Consistency of the DA and PoE strategies versus the consistency of DA and PoE for LLaMA-2 13B. Least consistent methods are \underline{underlined}.}
\end{table*}

\begin{table*}[t]
\centering
\small
\begin{tabular}{@{}ccccccc@{}}
 & \multicolumn{6}{c}{\emph{LLaMA-2 70B}} \\ \midrule
\multicolumn{1}{c|}{\multirow{2}{*}{Dataset}} & \multicolumn{3}{c|}{\textbf{Base}} & \multicolumn{3}{c}{\textbf{COT}}  \\
\multicolumn{1}{c|}{}  & DA Self & PoE Self & \multicolumn{1}{c|}{DA vs. PoE}  & DA Self & PoE Self & DA vs. PoE  \\ \midrule
\multicolumn{1}{c|}{Commonsense QA} & 97.3 & \underline{83.3} & \multicolumn{1}{c|}{86.0} & 94.4 & 83.0 & \underline{80.8} \\ 
\multicolumn{1}{c|}{Social IQa} & 95.2 & 90.9 & \multicolumn{1}{c|}{\underline{86.9}} & 90.0 & 85.9 & \underline{77.6} \\ \midrule
\multicolumn{1}{c|}{ARC} & 97.8 & 89.5 & \multicolumn{1}{c|}{\underline{77.8}} & 93.5 & 89.6 & \underline{85.3} \\ 
\multicolumn{1}{c|}{OpenBookQA} & 96.8 & 72.9 & \multicolumn{1}{c|}{\underline{67.6}} & 93.1 & 84.8 & \underline{81.4} \\ \bottomrule
\end{tabular}

\caption{\label{table:consistency_detailed_llama-2_70b}Self-Consistency of the DA and PoE strategies versus the consistency of DA and PoE for LLaMA-2 70B. Least consistent methods are \underline{underlined}.}
\end{table*}

\begin{table*}[t]
\centering
\small
\begin{tabular}{@{}ccccccc@{}}
 & \multicolumn{6}{c}{\emph{Falcon 7B}} \\ \midrule
\multicolumn{1}{c|}{\multirow{2}{*}{Dataset}} & \multicolumn{3}{c|}{\textbf{Base}} & \multicolumn{3}{c}{\textbf{COT}}  \\
\multicolumn{1}{c|}{}  & DA Self & PoE Self & \multicolumn{1}{c|}{DA vs. PoE}  & DA Self & PoE Self & DA vs. PoE  \\ \midrule
\multicolumn{1}{c|}{Commonsense QA} & 52.6 & 56.3 & \multicolumn{1}{c|}{\underline{50.6}} & 81.8 & 85.0 & \underline{25.6} \\ 
\multicolumn{1}{c|}{Social IQa} & 54.2 & 63.9 & \multicolumn{1}{c|}{\underline{50.8}} & 84.1 & 82.6 & \underline{27.2} \\ \midrule
\multicolumn{1}{c|}{ARC} & 58.7 & 50.3 & \multicolumn{1}{c|}{\underline{49.2}} & 78.1 & 74.6 & \underline{30.1} \\ 
\multicolumn{1}{c|}{OpenBookQA} & 55.5 & 59.1 & \multicolumn{1}{c|}{\underline{52.1}} & 82.2 & 76.9 & \underline{31.5} \\ \bottomrule
\end{tabular}

\caption{\label{table:consistency_detailed_falcon_7b}Self-Consistency of the DA and PoE strategies versus the consistency of DA and PoE for Falcon 7B. Least consistent methods are \underline{underlined}.}
\end{table*}

\begin{table*}[t]
\centering
\small
\begin{tabular}{@{}ccccccc@{}}
 & \multicolumn{6}{c}{\emph{Falcon 40B}} \\ \midrule
\multicolumn{1}{c|}{\multirow{2}{*}{Dataset}} & \multicolumn{3}{c|}{\textbf{Base}} & \multicolumn{3}{c}{\textbf{COT}}  \\
\multicolumn{1}{c|}{}  & DA Self & PoE Self & \multicolumn{1}{c|}{DA vs. PoE}  & DA Self & PoE Self & DA vs. PoE  \\ \midrule
\multicolumn{1}{c|}{Commonsense QA} & 97.3 & 84.8 & \multicolumn{1}{c|}{\underline{76.1}} & 91.4 & 67.6 & \underline{54.8} \\ 
\multicolumn{1}{c|}{Social IQa} & 94.5 & 88.0 & \multicolumn{1}{c|}{\underline{76.0}} & 86.2 & 75.0 & \underline{61.5} \\ \midrule
\multicolumn{1}{c|}{ARC} & 95.0 & 87.5 & \multicolumn{1}{c|}{\underline{68.1}} & 84.8 & 70.1 & \underline{50.3} \\ 
\multicolumn{1}{c|}{OpenBookQA} & 94.1 & 90.7 & \multicolumn{1}{c|}{\underline{73.6}} & 88.4 & 71.7 & \underline{61.9} \\ \bottomrule
\end{tabular}

\caption{\label{table:consistency_detailed_falcon_40b}Self-Consistency of the DA and PoE strategies versus the consistency of DA and PoE for Falcon 40B. Least consistent methods are \underline{underlined}.}
\end{table*}

\begin{table*}[t]
\centering
\small
\begin{tabular}{@{}ccccccc@{}}
 & \multicolumn{6}{c}{\emph{Falcon 180B}} \\ \midrule
\multicolumn{1}{c|}{\multirow{2}{*}{Dataset}} & \multicolumn{3}{c|}{\textbf{Base}} & \multicolumn{3}{c}{\textbf{COT}}  \\
\multicolumn{1}{c|}{}  & DA Self & PoE Self & \multicolumn{1}{c|}{DA vs. PoE}  & DA Self & PoE Self & DA vs. PoE  \\ \midrule
\multicolumn{1}{c|}{Commonsense QA} & 97.6 & 94.1 & \multicolumn{1}{c|}{\underline{91.4}} & 92.9 & 83.1 & \underline{78.2} \\ 
\multicolumn{1}{c|}{Social IQa} & 95.0 & 92.8 & \multicolumn{1}{c|}{\underline{86.6}} & 91.4 & 81.2 & \underline{69.4} \\ \midrule
\multicolumn{1}{c|}{ARC} & 98.5 & 94.1 & \multicolumn{1}{c|}{\underline{89.3}} & 90.8 & 88.3 & \underline{79.1} \\ 
\multicolumn{1}{c|}{OpenBookQA} & 96.2 & 94.4 & \multicolumn{1}{c|}{\underline{89.7}} & 95.1 & 83.1 & \underline{77.2} \\ \bottomrule
\end{tabular}

\caption{\label{table:consistency_detailed_falcon_180b}Self-Consistency of the DA and PoE strategies versus the consistency of DA and PoE for Falcon 180B. Least consistent methods are \underline{underlined}.}
\end{table*}

\begin{table*}[t]
\centering
\small
\begin{tabular}{@{}ccccccc@{}}
 & \multicolumn{6}{c}{\emph{GPT-3.5}} \\ \midrule
\multicolumn{1}{c|}{\multirow{2}{*}{Dataset}} & \multicolumn{3}{c|}{\textbf{Base}} & \multicolumn{3}{c}{\textbf{COT}}  \\
\multicolumn{1}{c|}{}  & DA Self & PoE Self & \multicolumn{1}{c|}{DA vs. PoE}  & DA Self & PoE Self & DA vs. PoE  \\ \midrule
\multicolumn{1}{c|}{Commonsense QA} & 97.3 & 89.2 & \multicolumn{1}{c|}{\underline{81.3}} & 97.6 & 93.3 & \underline{86.0} \\ 
\multicolumn{1}{c|}{Social IQa} & 97.0 & 91.6 & \multicolumn{1}{c|}{\underline{80.7}} & 95.2 & 89.1 & \underline{76.2} \\ \midrule
\multicolumn{1}{c|}{ARC} & 97.5 & 94.0 & \multicolumn{1}{c|}{\underline{86.6}} & 97.0 & 93.5 & \underline{85.5} \\ 
\multicolumn{1}{c|}{OpenBookQA} & 98.2 & 82.2 & \multicolumn{1}{c|}{\underline{72.3}} & 96.8 & 93.1 & \underline{81.4} \\ \bottomrule
\end{tabular}

\caption{\label{table:consistency_detailed_gpt-35}Self-Consistency of the DA and PoE strategies versus the consistency of DA and PoE for GPT-3.5. Least consistent methods are \underline{underlined}.}
\end{table*}

\clearpage

\begin{table*}[ht]
\tiny
\centering
\begin{tabularx}{\linewidth}{c | X | X}
\toprule
\textbf{Error Type} & \multicolumn{1}{c|}{\textbf{Question}} & \multicolumn{1}{c}{\textbf{Output}} \\ \midrule

Misaligned Rationale  & \specialcellleft{Question: Which biomolecule does \textcolor{blue}{\textbf{not have a carbon-nitrogen bond}}? \\
Choices: \\
(A) nucleic acid \\
(B) carbohydrate \\
Correct Answer: (B) carbohydrate} & \specialcellleft{Carbohydrates are made up of carbon, hydrogen, and oxygen.\\ \textcolor{red}{\textbf{There is no nitrogen in carbohydrates}}.\\ So the incorrect answer is "carbohydrate" which is choice (B)} \\ \midrule

Misaligned Rationale & \specialcellleft{Question: \textcolor{blue}{\textbf{What is someone}} who isnt clever, bright, or competent \textcolor{blue}{\textbf{called}}? \\
Choices: \\
(A) ineffectual \\
(B) stupid \\
Correct Answer: (B) stupid} & \specialcellleft{The term "stupid" \textcolor{red}{\textbf{is used to describe}} someone who is not \\ clever, bright, or competent. \\ So the incorrect answer is "stupid" which is choice (B)} \\ \midrule

Logical Error & \specialcellleft{Question: Which substance will retain the largest \\ amount of energy from the Sun? \\
Choices: \\
(A) sand \\
(B) grass \\
Correct Answer: (A) sand} & \specialcellleft{Sand is a poor conductor of heat, \textcolor{red}{\textbf{so it will not retain}} \\ \textcolor{red}{\textbf{much energy from the Sun}}. \\ So the incorrect answer is "sand" which is choice (A)} \\ \midrule

Logical Error & \specialcellleft{Question: Miss Grady took a stick from Bob because he was playing with it in\\class. She wanted to make sure that he couldn't get to it so she put it where? \\
Choices: \\
(A) desk drawer \\
(B) office \\
Correct Answer: (A) desk drawer} & \specialcellleft{Miss Grady wanted to make sure that Bob couldn't get to the stick, so she \\ would not have put it in the desk drawer \textcolor{red}{\textbf{where he could easily access it}}.\\So the incorrect answer is "desk drawer" which is choice (A)} \\ \midrule

No Breakdown & \specialcellleft{Question: Which object in the solar system is orbited by a belt of asteroids? \\
Choices: \\
(A) Saturn \\
(B) the Sun \\
Correct Answer: (B) the Sun} & \specialcellleft{\textcolor{red}{\textbf{The Sun is not orbited by a belt of asteroids}}. \\ So the incorrect answer is "the Sun" which is choice (B)} \\ \midrule

No Breakdown & \specialcellleft{Question: Where would you get some maps that you own? \\
Choices: \\
(A) cabinet \\
(B) important when traveling \\
Correct Answer: (A) cabinet} & \specialcellleft{\textcolor{red}{\textbf{Maps are important when traveling}}, \\ \textcolor{red}{\textbf{but you would not get them from a cabinet}}. \\ So the incorrect answer is "cabinet" which is choice (A)} \\ \midrule

Hallucination & \specialcellleft{Question: If a hyrdrogen atom gives up 1 electron, what is its overall charge? \\
Choices: \\
\textbf{\textcolor{blue}{\textbf{(A) +1}}} \\
(B) +2 \\
Correct Answer: (A) +1} & \specialcellleft{Atoms have a neutral charge, so if an atom gives up one electron, \\ it will have a positive charge. \\ So the incorrect answer is \textcolor{red}{\textbf{"neutral" which is choice (A)}}} \\ \midrule

Hallucination & \specialcellleft{Question: What do the feathers look like \textcolor{blue}{\textbf{on birds}} found in the rainforest? \\
Choices: \\
(A) singing \\
(B) very colorful \\
Correct Answer: (B) very colorful} & \specialcellleft{Feathers on birds found in the rainforest can be very colorful, \\ but not \textcolor{red}{\textbf{all birds}} found in the rainforest have colorful feathers. \\ So the incorrect answer is "very colorful" which is choice (B)} \\ \midrule

PoE N/A & \specialcellleft{Question: Arturo ran a 3,000-meter race. His running time from start \\ to finish was 10 minutes. What was Arturo's average speed? \\
Choices: \\
(A) 5 m/s \\
(B) 100 m/s \\
Correct Answer: (A) 5 m/s} & \specialcellleft{The question states that Arturo ran the race in 10 minutes, \\ but the question does not state the number of seconds in 10 minutes. \\ So the incorrect answer is "5 m/s" which is choice (A)} \\ \midrule

Dataset Issue & \specialcellleft{Question: When someone is physically competing what does their body do? \\
Choices: \\
\textcolor{blue}{\textbf{(A) releases heat}} \\
\textcolor{blue}{\textbf{(B) sweat}} \\
Correct Answer: (B) sweat} & \specialcellleft{When someone is physically competing, their body releases heat. \\ Sweat is a result of the body releasing heat. \\ So the incorrect answer is "sweat" which is choice (B)} \\ \bottomrule

\end{tabularx}
\caption{\label{appendix:error_table}Examples of errors in PoE COT rationales uncovered during the error analysis.}
\end{table*}

\begin{table*}[ht]
\tiny
\centering
\begin{tabularx}{\linewidth}{c | X | X }
\toprule
\textbf{Model} & \multicolumn{1}{c|}{\textbf{Question}} & \multicolumn{1}{c}{\textbf{Output}} \\ \midrule

LLaMA-2 70B  & \specialcellleft{Question: Which of these is a response of cats to external stimuli? \\
Choices: \\
(A) hairs on the back stand up when scared \\
(B) movement of waste into the bladder \\
Correct Answer: (A) hairs on the back stand up when scared} & \specialcellleft{The movement of waste into the bladder is an internal process, \\ so it is not a response to an external stimulus. So the incorrect answer \\ is "movement of waste into the bladder" which is choice (B)} \\ \midrule

LLaMA-2 70B  & \specialcellleft{Question: Where would someone keep their nylon leggings? \\
Choices: \\
(A) rope \\
(B) drawer \\
Correct Answer: (B) drawer} & \specialcellleft{Nylon leggings are a type of clothing and are not used to tie things together. \\So the incorrect answer is "rope" which is choice (A)} \\ \midrule

Falcon 180B  & \specialcellleft{Question: Which is a renewable natural resource that can be replenished \\over a period of time?\\
Choices:\\
(A) coal\\
(B) trees\\
Correct Answer: (B) trees} & \specialcellleft{Coal is a nonrenewable natural resource because it takes millions of years \\to form. So the incorrect answer is "coal" which is choice (A)} \\ \midrule

Falcon 180B  & \specialcellleft{Question: What do you do in order to fly in airplane? \\
Choices: \\
(A) buy tickets \\
(B) add gas \\
Correct Answer: (A) buy tickets} & \specialcellleft{Airplanes are large and require a lot of gas. \\ However, passengers do not add gas to the airplane. \\So the incorrect answer is "add gas" which is choice (B)} \\ \midrule

GPT 3.5  & \specialcellleft{Question: Which object occupies the greatest amount of space? \\
Choices: \\
(A) a galaxy\\
(B) a neutron star\\
Correct Answer: (A) a galaxy} & \specialcellleft{A neutron star is much smaller and denser than a galaxy,\\ so it does not occupy the greatest amount of space.\\ So the incorrect answer is "a neutron star" which is choice (B)} \\ \midrule

GPT 3.5  & \specialcellleft{Question: Anybody could be hired in the kitchen, what was needed of them? \\
Choices: \\
(A) forget\\
(B) wash dishes\\
Correct Answer: (B) wash dishes} & \specialcellleft{If someone is hired in the kitchen, they are expected to work and perform tasks.\\ Forgetting is not a desirable trait or skill in a kitchen.\\ So the incorrect answer is "forget" which is choice (A)} \\ \bottomrule

\end{tabularx}
\caption{\label{appendix:success_table}Examples of sound rationales elicited from PoE COT.}
\end{table*}

\clearpage
  
\begin{table*}[t]\
\centering
\small
\begin{adjustbox}{width=1\textwidth} 
\begin{tabular}{@{}cl|ll|ll|ll@{}}\toprule\multicolumn{2}{c|}{\multirow{2}{*}{Metrics}} & \multicolumn{2}{c|}{\textbf{LLaMA 7b}} &  \multicolumn{ 2}{c|}{\textbf{LLaMA 13b}} &  \multicolumn{ 2}{c}{\textbf{LLaMA 70b}}\\ \multicolumn{2}{c|}{} & Base & COT & Base & COT & Base & COT \\ \midrule 

\multirow{3}{*}{5 Choices} & Raw Accuracy & 0.530& 0.664& 0.646& 0.69& 0.814& 0.782 \\ & DA Accuracy & 0.530& 0.665& 0.646& 0.691& 0.814& 0.784 \\ & PoE Accuracy & 0.594& 0.680& 0.456& 0.782& 0.942& 0.940 \\ \midrule

\multirow{3}{*}{4 Choices} & Raw Accuracy & 0.322& 0.450& 0.262& 0.580& 0.782& 0.774 \\ & DA Accuracy & 0.542& 0.662& 0.575& 0.742& 0.830& 0.823 \\ & PoE Accuracy & 0.362& 0.486& 0.230& 0.526& 0.902& 0.864 \\ \midrule

\multirow{3}{*}{3 Choices} & Raw Accuracy & 0.232& 0.350& 0.158& 0.416& 0.778& 0.710 \\ & DA Accuracy & 0.641& 0.720& 0.687& 0.791& 0.863& 0.826 \\ & PoE Accuracy & 0.224& 0.348& 0.128& 0.376& 0.788& 0.760 \\ \midrule 

\multirow{3}{*}{2 Choices} & Raw Accuracy & 0.168& 0.270& 0.092& 0.330& 0.706& 0.662 \\ & DA Accuracy & 0.750& 0.776& 0.719& 0.887& 0.896& 0.873 \\ & PoE Accuracy & 0.084& 0.194& 0.056& 0.210& 0.636& 0.598 \\ \bottomrule 

\end{tabular}\end{adjustbox}\caption{Iterative process of elimination results on Commonsense QA. Raw accuracy denotes the proportion of all questions that are answered correctly. DA Accuracy and PoE Accuracy are the accuracies of the two strategies, accounting for the errors that could have previously been made by PoE. DA and PoE Accuracy are computed as the accuracy conditioned on if the remaining choices still contain the gold answer.}\label{appendix:cqa_iterative}

\end{table*}

\begin{table*}[t]
\centering
\small
\begin{adjustbox}{width=1\textwidth} 
\begin{tabular}{@{}cl|ll|ll|ll@{}}\toprule\multicolumn{2}{c|}{\multirow{2}{*}{Metrics}} & \multicolumn{ 2}{c|}{\textbf{LLaMA 7b}} &  \multicolumn{ 2}{c|}{\textbf{LLaMA 13b}} &  \multicolumn{ 2}{c}{\textbf{LLaMA 70b}}\\

\multicolumn{2}{c|}{} & Normal & COT & Normal & COT & Normal & COT \\ \midrule

\multirow{3}{*}{4 Choices} & Raw Accuracy & 0.526& 0.596& 0.642& 0.664& 0.834& 0.814 \\ & DA Accuracy & 0.530& 0.601& 0.647& 0.671& 0.844& 0.821 \\ & PoE Accuracy & 0.572& 0.642& 0.458& 0.634& 0.926& 0.892 \\ \midrule

\multirow{3}{*}{3 Choices} & Raw Accuracy & 0.322& 0.422& 0.278& 0.426& 0.812& 0.750 \\ & DA Accuracy & 0.563& 0.659& 0.61& 0.681& 0.877& 0.854 \\ & PoE Accuracy & 0.334& 0.380& 0.238& 0.392& 0.832& 0.808 \\ \midrule

\multirow{3}{*}{2 Choices} & Raw Accuracy & 0.232& 0.288& 0.176& 0.306& 0.768& 0.744 \\ & DA Accuracy & 0.699& 0.766& 0.752& 0.797& 0.923& 0.937 \\ & PoE Accuracy & 0.150& 0.216& 0.106& 0.222& 0.624& 0.708 \\ \bottomrule

\end{tabular}
\end{adjustbox}
\caption{Iterative process of elimination results on ARC. Raw accuracy denotes the proportion of all questions that are answered correctly. DA Accuracy and PoE Accuracy are the accuracies of the two strategies, accounting for the errors that could have previously been made by PoE. DA and PoE Accuracy are computed as the accuracy conditioned on if the remaining choices still contain the gold answer.}
\label{appendix:arc_iterative}
\end{table*}

\clearpage

\end{document}